\documentclass[runningheads, envcountsame, a4paper]{llncs}
\usepackage{amsmath,amssymb, graphicx}
\usepackage[ruled,linesnumbered]{algorithm2e}
\usepackage{algorithmic}
\usepackage{paralist} % also compact lists
\usepackage{color}
\usepackage{multirow}
\usepackage{rotating}
\usepackage{comment}
\usepackage{grffile}
\usepackage[mathscr]{eucal} 
\usepackage{amsbsy} 
 
\usepackage{subcaption}
\usepackage{booktabs}
\usepackage{xcolor}
\usepackage{tikz}
\usetikzlibrary{snakes}
\usepackage{multirow}
\usepackage{mdframed,lipsum}
\usepackage{epstopdf}
\usepackage{balance}
\usepackage{tablefootnote}
\usepackage{empheq} % added by christos - makes nice box for equations
\usepackage{moresize}
\usepackage{enumitem}
\usepackage{xcolor}
\usepackage{optidef}
\usepackage[misc]{ifsym}
\usepackage[strings]{underscore}
\usepackage[hyphens]{url}

\setlist{nolistsep}

\newcommand{\hide}[1]{}

\graphicspath{{figs/}}

\begin{document}
% \title{Sparse Subspace Clustering analysis of Neural Networks}
 \title{Subspace Clustering Based Analysis of Neural Networks}
 
 \author{Uday Singh Saini \Letter \inst{1} \and
Pravallika Devineni \inst{2} \and
Evangelos E. Papalexakis \inst{1}}
\authorrunning{U.S. Saini, P. Devineni, E.E. Papalexakis}
\titlerunning{Subspace Clustering Based Analysis of Neural Networks}
\toctitle{Subspace Clustering Based Analysis of Neural Networks}
\tocauthor{Uday Singh Saini, Pravallika Devineni, Evangelos E. Papalexakis}
% First names are abbreviated in the running head.
% If there are more than two authors, 'et al.' is used.
%
\institute{University of California Riverside, 900 University Avenue, Riverside, CA, USA \\
\email{usain001@ucr.edu, epapalex@cs.ucr.edu}\\
%\url{http://www.springer.com/gp/computer-science/lncs} \and
\and
Oak Ridge National Laboratory, Oak Ridge, TN, USA \\
\email{devinenip@ornl.gov}}

\maketitle
\setcounter{footnote}{0}

\begin{abstract}
Tools to analyze the latent space of deep neural networks provide a step towards better understanding them. In this work, we motivate sparse subspace clustering (SSC) with an aim to learn affinity graphs from the latent structure of a given neural network layer trained over a set of inputs. We then use tools from Community Detection to quantify structures present in the input. These experiments reveal that as we go deeper in a network, inputs tend to have an increasing affinity to other inputs of the same class. Subsequently, we utilise matrix similarity measures to perform layer-wise comparisons between affinity graphs. In doing so we first demonstrate that when comparing a given layer currently under training to its final state, the shallower the layer of the network, the quicker it is to converge than the deeper layers. When performing a pairwise analysis of the entire network architecture, we observe that, as the network increases in size, it reorganises from a state where each layer is moderately similar to its neighbours, to a state where layers within a block have high similarity than to layers in other blocks. Finally, we analyze the learned affinity graphs of the final convolutional layer of the network and demonstrate how an input's local neighbourhood affects its classification by the network. 
\end{abstract}

\section{Introduction}
\label{sec:intro}
With the emergence of deep neural networks in a variety of domains, there is a need for understanding and characterising the inner workings and operations of these models. With critical applications such as autonomous vehicles, healthcare, and criminal justice relying on neural network-based models, model interpretability has become an important and necessary aspect of machine learning. In the domain of theoretical analysis of neural networks, previous works like \cite{lin2018resnet}, \cite{hanin2018approximating}, and \cite{lu2017expressive} focus on the approximation capability of neural networks, while works like \cite{lee2018deep} and \cite{novak2020bayesian} focus on the interpretations of neural networks as kernels. In the domain of experimental analysis, works like \cite{zeiler2013visualizing}, \cite{simonyan2014deep} and \cite{bau2017network} focus on visualising inputs, filters and neurons of a network. 

Our work focuses on interpreting the behaviour of neural networks by analyzing subspaces of latent representations learned by the layers of a neural network. Most closely related to our approach are works like SVCCA \cite{raghu2017svcca}, PWCCA \cite{morcos2018insights} and Linear-CKA \cite{kornblith2019similarity}, \cite{nguyen2020wide}. SVCCA and PWCCA operate directly on the activations of a neural network for a given set of inputs and allows comparison between two such sources of activations over the same inputs, thereby facilitating a comparison between different layers of the same network, or even different networks. On the other hand, \cite{kornblith2019similarity} and \cite{nguyen2020wide} take the same activations over a set of inputs and use it to construct a pairwise similarity matrix with the help of a kernel. Taking these pairwise similarity matrices obtained via the response of two different layers to a set of inputs, they then compare the two kernel matrices using Centered Kernel Alignment (CKA) \cite{DBLP:journals/corr/abs-1203-0550}. This lets them directly compare any two layers from different sources. For most of their experiments, \cite{kornblith2019similarity} and \cite{nguyen2020wide} motivate the theory and use of linear kernels, henceforth we call their method Linear-CKA. All these methods offer architecture agnostic ways to analyze neural networks.

We investigate Sparse Subspace Clustering (SSC) \cite{elhamifar2013sparse} as an alternative to Kernel-based pairwise Similarity matrices over inputs. Methods like SSC help us learn a connectivity graph over data points that are assumed to belong to an underlying union of subspaces embedded in a given space. Such methods are typically based on the \textbf{Self-Expressiveness} property in a union of subspaces, where each data point can be represented as a sparse linear combination of points from its own subspace. Learning such a graph helps impart transparency to the learning dynamics of the neural network and also presents us with a way to represent each layer of a neural network in an architecture agnostic manner. This facilitates the comparison of layers within a network and across various architectures. 

We combine SSC with CKA to provide an alternate similarity measurement tool for latent representations learned by neural networks. It is akin to translating a layer's activations over a set of inputs into a connectivity graph and then using CKA to compute the similarity between graphs arising from two different neural network representations.

Our main contributions are the following: 
\begin{enumerate}
    \item We utilise SSC to learn a connectivity graph over inputs in the latent space. We then employ graph modularity \cite{Newman8577} to characterise the community structure of each input's neighbourhood. We demonstrate that as we go deeper into the layers of the network, the community around each input becomes more homogeneous, i.e., a higher proportion of an input's neighbourhood in the Affinity Graph is of the same class as the input itself. Additionally, we also demonstrate the utility of SSC-CKA to capture network training dynamics. We observe that when compared to shallower layers, the deeper layers of the network tend to take longer to converge to their final state. This is a corroboration of similar observations made in SVCCA \cite{raghu2017svcca}.    
    \item We demonstrate the ability of SSC-CKA to visualize and analyze entire network architectures by capitalising on its ability to perform pairwise comparison of two different layers of a network (or two different networks). These experiments set along the lines of Linear-CKA \cite{kornblith2019similarity}, demonstrate the effects of depth, width, epochs and training data size on the network architecture.
    \item We demonstrate the ability of SSC to interpret the latent spaces of a given layer of the network. SSC allows us to represent each input as a weighted and sparse linear combination of other inputs in the same subspace. Here we demonstrate a strong correlation between classes of network's prediction for the input and the class having the highest weightage in the local neighbourhood of the input. This hints towards neural networks separating out classes into disjoint subspaces and presents interesting opportunities and directions going forward for creating tools to analyze neural networks. 
\end{enumerate}

\section{Background and Method}
\label{sec:method}

In this section, we lay the background on the Sparse Subspace Clustering, its optimization through ADMM \cite{10.1561/2200000016} and motivate its fitness for the purpose of interpreting subspaces learned by neural networks.

\subsection{Sparse Spectral Clustering}
Subspace clustering refers to the task of clustering the data into their original subspaces and uncovering the
underlying structure of the data. Given some high-dimensional data, subspace clustering attempts to model the data as samples drawn from a union of multiple low-dimensional linear subspaces \cite{elhamifar2013sparse}. First, we present Sparse Spectral Clustering (SSC) for the case of uncorrupted data. As mentioned earlier, SSC typically works well when the underlying data follows Self-Expressive Property, i.e., each data point can be represented by a linear combination of points that belong to the same subspace. 

Let $X$ = $[\mathbf{x}_1,\dots,\mathbf{x}_N]$, such that $X \in \mathbf{R}^{d \times N}$ represents the data matrix. Let $C$ = $[\mathbf{c}_1,\dots,\mathbf{c}_N]$, such that $C \in \mathbf{R}^{N \times N}$ and $\mathbf{c}_i \in \mathbf{R}^{N}$, where the entry $(i,j)$ of the matrix $C$, given by $c_{ij}$ represents the weight of data point $\mathbf{x}_j \in \mathbf{R}^{d}$ in the linear combination to reconstruct $\mathbf{x}_i \in \mathbf{R}^{d}$. Mathematically, a noiseless model for SSC is equivalent to Equation \ref{eq:SSC}. However, as shown in \cite{10.1016/S0304-3975(97)00115-1}, this problem is NP-hard.

% \begin{align*}
	\begin{equation}
	\begin{aligned}
	\label{eq:SSC}
	    \min_{\mathbf{c}_i} ||\mathbf{c}_i||_{0} \quad \textnormal{s.t.} \quad \mathbf{x}_i = X\mathbf{c}_i \textnormal{, }  c_{ii} = 0  \quad \forall i \in \{1,\dots,N\}
	   % \caption{Objective Function: Sparse Subspace Clustering - Noiseless}
	   \end{aligned}
	\end{equation}
% 	\caption{Objective Function: Sparse Subspace Clustering - Noiseless}
% \end{align*}

 We therefore focus on Equation \ref{eq:SSCL1}, a convex relaxation of Equation \ref{eq:SSC} which is also robust to noise and solve this optimization problem using Algorithm \ref{algo:SSCADMM} as shown in \cite{10.5555/2948884}.
\begin{equation}
% % 	\begin{equation}
    \begin{aligned}
	\label{eq:SSCL1}
	%\underset{C,Z}{\text{minimize  }} ||C||_{1} + \frac{\tau}{2} ||X - XZ||^{2}_{F} ~ \ni Z = C - diag(C)
	\min_{C,Z} ||C||_{1} + \frac{\tau}{2} ||X - XZ||^{2}_{F} \quad \textnormal{s.t.} \quad  Z = C - diag(C)
	\end{aligned}
% % 	\end{equation}
% 	\caption{Objective Function: Sparse Subspace Clustering with Noise}
\end{equation}

\subsection{Centered Kernel Alignment}
Centered Kernel Alignment \cite{DBLP:journals/corr/abs-1203-0550}, as defined in Equation \ref{eq:CKA}, between two similarity matrices $X$ and $Y$ is an isotropic invariant similarity index that relies upon Hilbert-Schmidt Independence Criterion (HSIC) \cite{10.1007/11564089_7} to determine statistical dependence between two sets of variables.
\begin{equation}
	\label{eq:CKA}
	    CKA(X,Y) = \frac{HSIC(X,Y)}{\sqrt{HSIC(X,X)HSIC(Y,Y)}}
	   % \caption{Objective Function: Sparse Subspace Clustering - Noiseless}
	\end{equation}

where HSIC between a pair of $N \times N$ matrices is defined in Equation \ref{eq:HSIC}
\begin{equation}
	\label{eq:HSIC}
	    HSIC(X,Y) = \frac{trace(HXHHYH)}{(N-1)^{2}}
	   % \caption{Objective Function: Sparse Subspace Clustering - Noiseless}
	\end{equation}

where H is a Centering matrix given by $H = I - \frac{1}{N}\textbf{11}^{T}$.

\subsection{Meta Algorithm}
Our goal is to take a matrix of neural activations $X \in \mathbf{R}^{d_{1} \times N}$, where $d_1$ is the number of neurons in the given layer and $N$ is the number of examples for which we obtain the activations, and construct an affinity matrix $C_{X} \in \mathbf{R}^{N \times N}$ , via Sparse Subspace Clustering \cite{elhamifar2013sparse} , where each non diagonal entry $(i,j)$ of $C_{X}$ denotes the affinity between input samples $i$ and $j$. This gives us the ability to compare 2 matrices of neural activations, say $X \in \mathbf{R}^{d_{1} \times N}$ and $Y \in \mathbf{R}^{d_{2} \times N}$ by representing them as 2 Affinity Matrices $\in$ $\mathbf{R}^{N \times N}$, namely $C_{X}$ and $C_{Y}$ and then using Centered Kernel Alignment\cite{DBLP:journals/corr/abs-1203-0550} to compare the similarity of the 2 Affinity Matrices. This procedure helps us compare 2 different layers of a network, or 2 different layers of 2 architecturally different neural networks. We would like to point out that $C_{X} = |C| + |C^{T}|$ where the matrix $C$ is obtained from Algorithm \ref{algo:SSCADMM}, where $|C|$ represents the absolute value function applied element-wise to $C$.

\begin{algorithm}
    \SetAlgoLined
    \KwData{Data Matrix $X$}
    \KwResult{Sparse Representation $C$}
     \textbf{initialization:} $C^0 = \textbf{0}$, $\Lambda^{0}_{2} = \textbf{0}$, $\mu_{2} > 0$\;
     \While{not converged}{
      $Z^{k+1} = (\tau X^TX + \mu_{2}I)^{-1}(\tau X^TX + \mu_{2}(C^{k} - \frac{\Lambda^{k}_{2}}{\mu_{2}} ))$\;
      $C^{k+1} = \textbf{S}_{\frac{1}{\mu_{2}}}(Z^{k+1} +  \frac{\Lambda^{k}_{2}}{\mu_{2}})$; \textbf{S} : Shrinkage operator\
      $C^{k+1} = C^{K+1} - diag(C^{k+1})$\;
      $\Lambda^{k+1}_{2} = \Lambda^{k}_{2} + \mu_{2}(Z^{K+1} - C^{k+1})$\;

     }
     \caption{Matrix LASSO Minimization by ADMM}
     \label{algo:SSCADMM}
\end{algorithm}

\section{Problem and Experimental Setup}
\label{sec:problem}
In this paper, we experiment with VGGs \cite{simonyan2015deep}, ResNets \cite{he2015deep}, Wide-ResNets \cite{zagoruyko2017wide} and DenseNets \cite{huang2018densely} trained on CIFAR-10 and CIFAR-100 datasets \cite{Krizhevsky09learningmultiple}. Our approach relies upon having access to activations of various hidden layers in the network for each input instance. As an example, for the first set of experiments in this study, we take outputs from a few pooling layers at the end of each block of a DenseNet to study the community structure of its subspace and also to learn the layer-wise dynamics of training progression. In another setting, we take focus on all the convolutional layers in the  network and get their activations for an input.  
\subsection{Problem Formulation}
We attempt to interpret and analyze neural networks by taking a matrix of its activations (layer-wise) $X \in \mathbf{R}^{d \times N}$, where $d$ is the number of neurons in the subject layer and $N$ is the number of inputs used for analysis. We first aim to learn a connectivity graph between these inputs based on their affinity scores obtained from SSC. Then we use this connectivity matrix to perform analysis pertaining to the community structure of the graph in Section \ref{sec:experiments}, and in Section \ref{sec:resultsb} we analyze instance based  neighbourhoods of various inputs to better reconcile network predictions with the need to human oriented explanations  
\subsection{Experimental Details}
All networks, unless otherwise stated, were trained with a learning rate of 0.1 and a weight decay of $5 \times 10^{-4}$ with a learning rate step multiplier of 0.2 applied after every 30 epochs. All networks were trained for 100 epochs since that was sufficient to achieve optimal performance, with the exception of VGG-29, which required 160 epochs. For SSC computations, both $\tau$ and $\mu$ were set to 10, with $\mu$ being adaptive based on equation 3.13 in \cite{10.1561/2200000016}. We recommend choosing $\tau$ and $\mu$ between 10 and 100. Our implementation is publicly available on GitHub \footnote{URL - \url{ https://github.com/23Uday/Subspace-Clustering-based-analysis-of-Neural-Networks}}.

\section{Analysis of Network Training Dynamics}
\label{sec:experiments}

In this section, we analyze behaviour of the network as its training progresses. To demonstrate the training dynamics, we train the following networks\footnote{Networks used from: \url{https://github.com/kuangliu/pytorch-cifar} and \url{https://github.com/meliketoy/wide-resnet.pytorch}} -  ResNet \cite{he2015deep}, Wide ResNet \cite{zagoruyko2017wide} and DenseNet \cite{huang2018densely} on the datasets CIFAR-10 and CIFAR-100 \cite{Krizhevsky09learningmultiple}. Both ResNets and DenseNets are constructed by stacking and joining different residual blocks with multiple convolution layers residing in each block. For brevity and scalability of this experiment, we use the output of residual blocks instead of every convolution layer and the final classification layers. For a given combination of the network and the dataset, we present three results - layer-wise modularity of the sparse subspace affinity graphs, SSC-CKA based layer-wise similarity to understand training dynamics, and analogous training dynamic analysis with Linear-CKA.

\subsection{Community Structure via Graph Modularity}

We analyze the community structure of the sparse subspace affinity graph for various layers at each epoch. We do this by calculating the modularity \cite{Newman8577} of the learned subspace affinity graph. Modularity is a measure of the structure of a graph, measuring the density of connections within a module or community. High modularity in a graph indicates dense connectivity between nodes of the same type, while low modularity indicates dense connectivity between nodes of different types, where a node type is its class label.  Mathematically, modularity of a graph can be represented as follows

\begin{equation}
\label{eq:Modularity}
    Q = \frac{1}{2m}\sum_{ij}\left( A_{ij} - \frac{k_i k_j}{2m}\right)\delta(c_i,c_j)
    % \caption{Definition: Modularity of a Graph}
\end{equation}

where $m$ is the number of edges in the graph, $A_{ij}$ is the weight of the edge between nodes $i$ and $j$, $k_i$  is the degree of node $i$, $c_i$ is the class label of node $i$ and $\delta(c_i,c_j)$ is 1 if node $i$ and node $j$ are of the same class and 0 otherwise.

We present this analysis in \iffalse Figure \ref{fig:res18c10a}, Figure \ref{fig:res18c100a}, Figure \ref{fig:wres24c10a}, Figure \ref{fig:wres24c100a},\fi Figure \ref{fig:den121c10a} and Figure \ref{fig:den121c100a}. We observe that as we go deeper in the network, the modularity of the learned sparse subspace affinity graph increases, implying that earlier layers of the network cluster examples of different classes together in the same subspace and deeper layers of the network tend to separate out classes into different disjoint subspaces. Another noteworthy observation is that the modularity scores of subspace affinity graph learned from earlier layers tends to saturate earlier in training when compared to the modularity of the subspace affinity graphs of deeper layers. This phenomenon is consistent across different architectures and different datasets. A similar observation regarding earlier saturation of shallower layers in training dynamics is also made in \cite{raghu2017svcca}, albeit in context of representational similarity, which we address next.

\subsection{Layer-wise Training Dynamics and Convergence}
For our second analysis, we compute the Centered Kernel Alignment (CKA) scores between the sparse subspace affinity graph of a layer at a given epoch and the same layer after the final epoch. This allows us to observe the rates at which layers converge to their final states. These results are shown in \iffalse Figure \ref{fig:res18c10b}, Figure \ref{fig:res18c100b}, Figure \ref{fig:wres24c10b}, Figure \ref{fig:wres24c100b},\fi Figure \ref{fig:den121c10b} and Figure \ref{fig:den121c100b}. We observe a similar pattern of shallower layers converging to their final representation much earlier, when compared to deeper layers of the network. Another key observation that we make pertains to the behaviour of representations when the learning rate of the optimiser is reduced to improve convergence. For training of all the networks, we start with a learning rate of 0.1 and reduce it by a factor of 0.2 after every 30 epochs using SGD with learning rate decay. At each step size decay, in addition to an improvement in network accuracy, we also observe a jump in the CKA similarity scores of the epoch's subspace affinity graph towards the CKA scores of final epoch's subspace affinity graph. We note that this jump is less prominent in shallower layers of the network when compared to the deeper layers as shallower layers converge to their final representations much earlier in training when compared to deeper layers. This bottom-up convergence observation is similar to observations made in \cite{morcos2018insights}, and \cite{raghu2017svcca} as described earlier. This jump also signifies a marked deviation from the representations of the layer during it's previous state - before the step decay, and current state - after the step decay. These observations highlight the role of step decay in Stochastic Gradient Descent based training of Neural networks to escape saddle points and other spurious local minima.

\subsection{Comparison with Linear-CKA:}
\begin{figure*}[!h]
 \begin{subfigure}{0.32\textwidth}
    \centering
    \includegraphics[width=\linewidth]{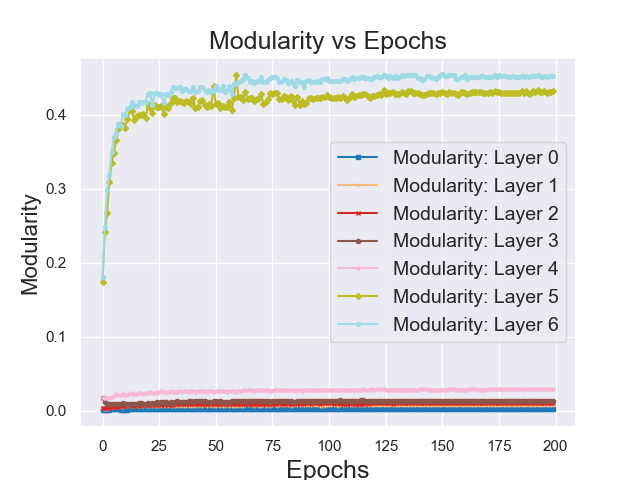}
    \caption{Modularity vs Epochs}
    % \caption{Modularity w.r.t. Epochs of the SSC-Affinity Graph of a DenseNet-121 on CIFAR-10 Dataset.}
    \label{fig:den121c10a}
  \end{subfigure}
    \begin{subfigure}{0.32\textwidth}
    \centering
    \includegraphics[width=\linewidth]{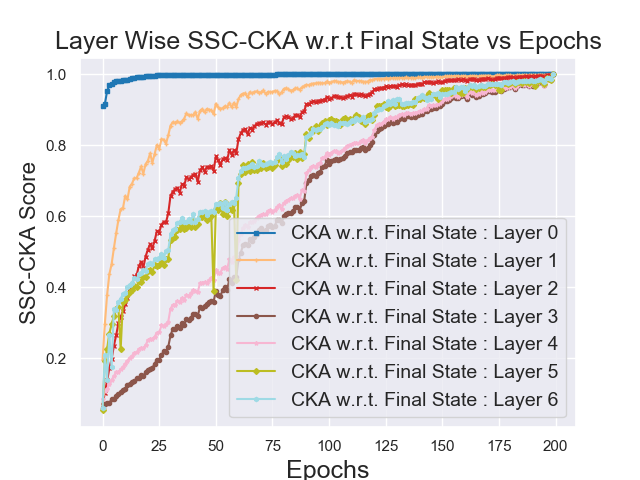}
    \caption{ SSC-CKA  w.r.t. Final State}
    % \caption{ CKA of the SSC-Affinity Graph of a DenseNet-121 w.r.t. the Final State  on CIFAR-10 Dataset.}
    \label{fig:den121c10b}
  \end{subfigure}
  \begin{subfigure}{0.32\textwidth}
        \centering
        \includegraphics[width=\linewidth]{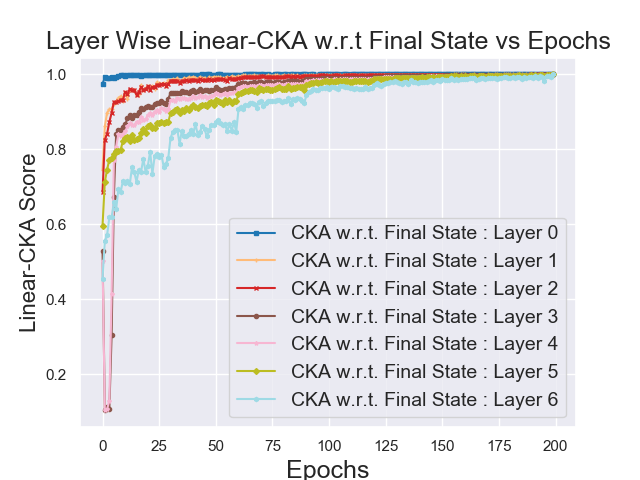}
        \caption{ Linear-CKA  w.r.t. Final State}
        % \caption{ CKA of the Pairwise dot Product of a DenseNet-121 w.r.t. the Final State  on CIFAR-10 Dataset.}
        \label{fig:den121c10c}
      \end{subfigure}
      
      \caption{Analysis of SSC Affinity Graphs using Modularity and CKA and it's comparison with a Linear Kernel Affinity Graph on DenseNet-121 network using  CIFAR-10 dataset.}
      \label{fig:den121c10}
\end{figure*}

\begin{figure*}[!t]
 \begin{subfigure}{0.32\textwidth}
    \centering
    \includegraphics[width=\linewidth]{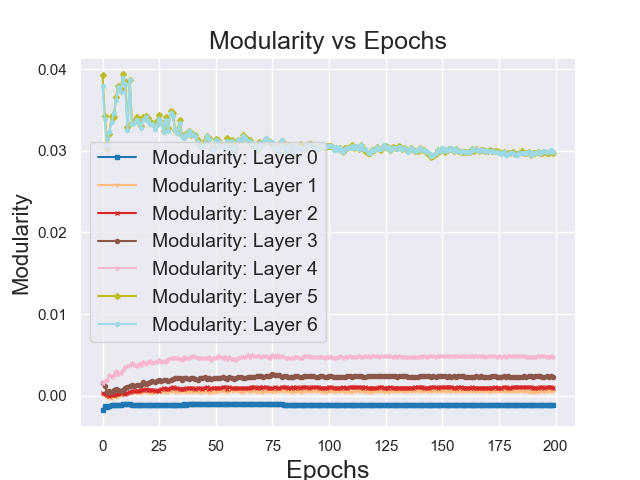}
    \caption{Modularity vs Epochs}
    % \caption{Modularity w.r.t. Epochs of the SSC-Affinity Graph of a DenseNet-121 on CIFAR-100 Dataset.}
    \label{fig:den121c100a}
  \end{subfigure}
    \begin{subfigure}{0.32\textwidth}
    \centering
    \includegraphics[width=\linewidth]{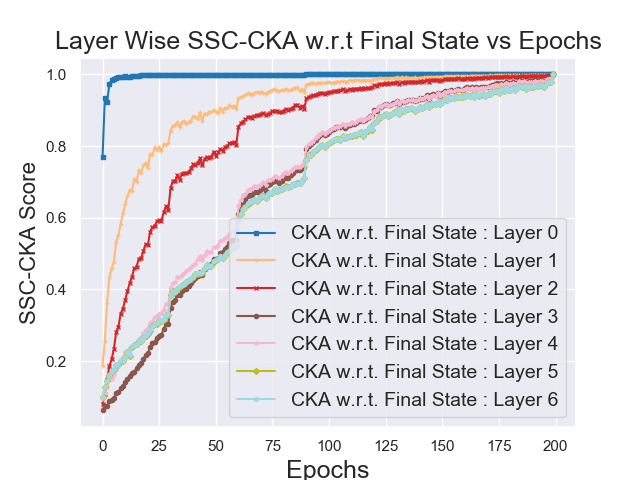}
    \caption{ SSC-CKA  w.r.t. Final State}
    % \caption{ CKA of the SSC-Affinity Graph of a DenseNet-121 w.r.t. the Final State  on CIFAR-100 Dataset.}
    \label{fig:den121c100b}
  \end{subfigure}
  \begin{subfigure}{0.32\textwidth}
        \centering
        \includegraphics[width=\linewidth]{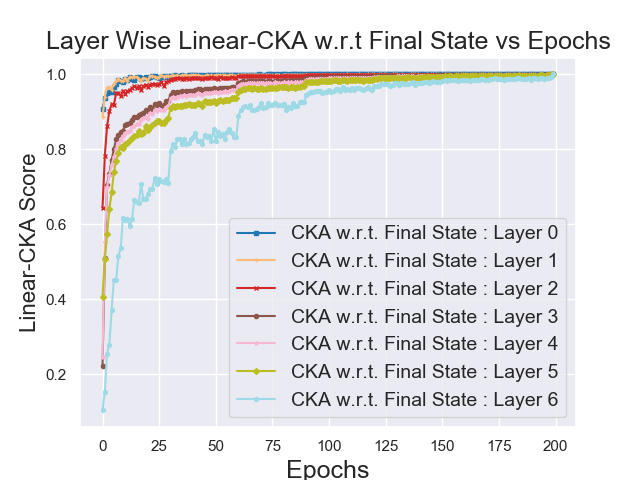}
        \caption{ Linear-CKA  w.r.t. Final State}
        % \caption{ CKA of the Pairwise dot Product of a DenseNet-121 w.r.t. the Final State  on CIFAR-100 Dataset.}
        \label{fig:den121c100c}
      \end{subfigure}
      
      \caption{Analysis of SSC Affinity Graphs using Modularity and CKA and it's comparison with a Linear Kernel Affinity Graph on DenseNet-121 network using  CIFAR-100 dataset.}
      \label{fig:den121c100}
\end{figure*}

As a comparison with related works in the area in \iffalse Figure \ref{fig:res18c10c},  Figure \ref{fig:res18c100c},  Figure \ref{fig:wres24c10c},  Figure \ref{fig:wres24c100c},\fi  Figure \ref{fig:den121c10c} and  Figure \ref{fig:den121c100c} we present the same training dynamics as previous paragraph, this time by analyzing the Linear Kernel-CKA \cite{kornblith2019similarity}. We observe a much cleaner bottom-up convergence of layers when compared with our method SSC-CKA, but the rates of convergence to final state is much faster when compared with the rates from SSC-CKA.

% \vspace{-.7cm}
% \end{comment}

\section{Analysis of Network Architecture}
\label{sec:results}
% \subsection{Network Architecture Analysis}
% \vspace{-.7cm}
In this section, we utilise SSC-CKA to visualize network architectures. We do this by using CKA to compare the Subspace Affinity Graphs obtained by applying SSC on the activations of two different layers of the network.  We try to focus on various inter-layer dynamics in different scenarios so as to learn intrinsic behaviours of the network and the architecture class it belongs to. In pursuit of these goals, we evaluate and demonstrate the architectural behaviour of various networks as a function of network architectural depth, network width, training duration in terms of epochs and finally as a function of quantity of training data. The motivation behind these experiments is to decipher how the latent space structure evolves through different layers of the network and how does a given set of layers add to the modeling power of the network. In doing so, we observe that deeper networks, wider networks, prematurely trained networks (networks trained for fewer epochs than optimal) and malnourished networks (networks trained on less training data but trained till saturation of performance) tend to develop prominent and mostly non-overlapping block diagonal structures in their layer-wise SSC-CKA heatmaps. This indicates that as the network's modeling capacity increases, a given amount training data is unable to exhaust the spare modeling bandwidth available due to additional layers, thus, a network tends to reorganise itself into blocks of layers with a high intra-block similarity among layers and a low inter-block similarity between layers.  The networks used in these set of experiments are VGGs \cite{simonyan2015deep}, ResNets \cite{he2015deep}, Wide-ResNets \cite{zagoruyko2017wide}, all experiments conducted with CIFAR-10 \cite{Krizhevsky09learningmultiple} as the dataset.

\subsection{Observing the Effects of Depth}
\begin{figure*}[!ht]
    \label{fig:vgg29c10}
    \begin{subfigure}{.24\textwidth}
        \centering
        \includegraphics[width=\linewidth]{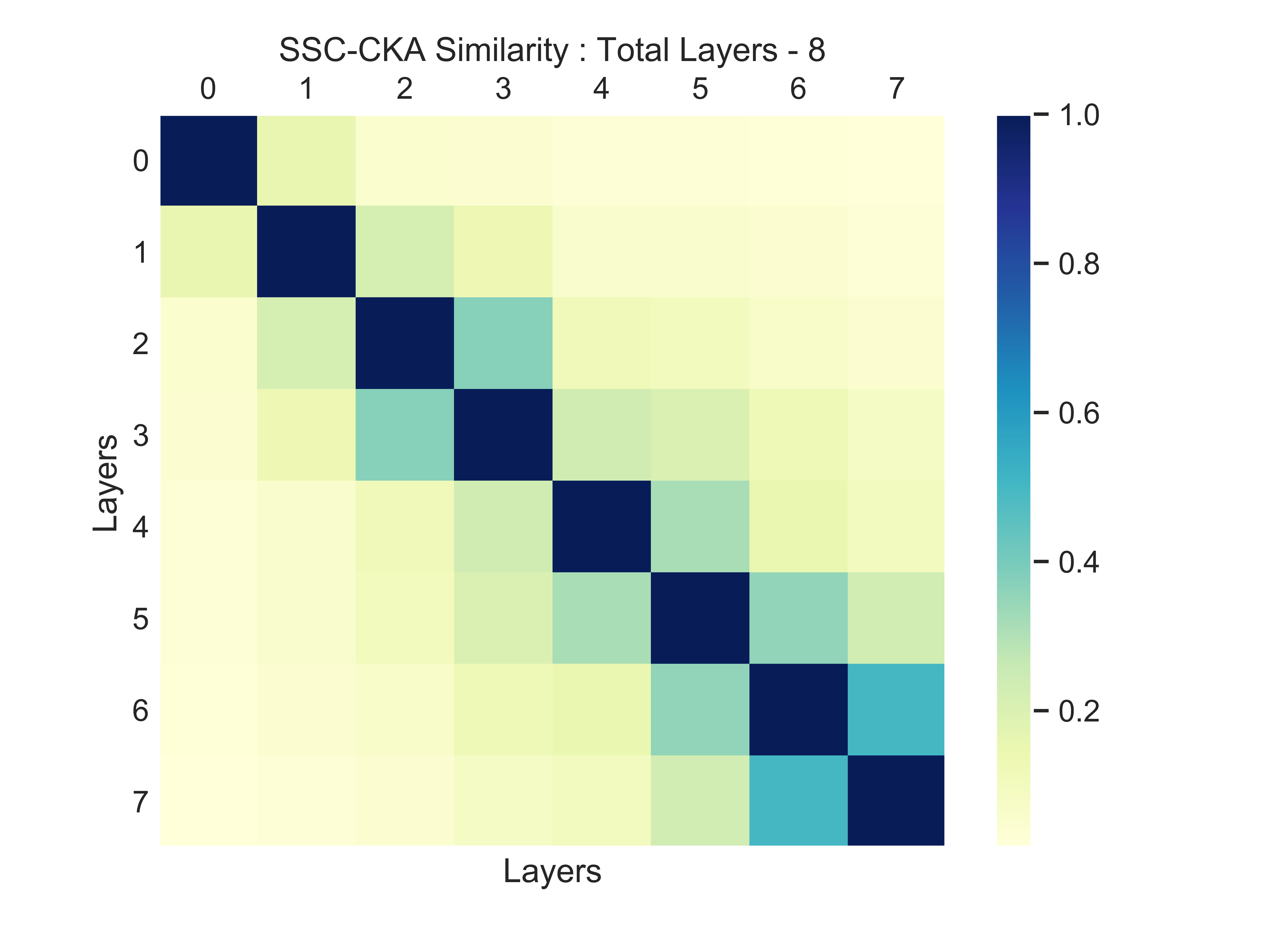}
        \caption{VGG-11}
        \label{fig:vgg11c10a}
      \end{subfigure}
    %     \begin{subfigure}{.32\textwidth}
    %     \centering
    %     \includegraphics[width=\linewidth]{LayerWiseAnalysis/Cifar10/VGG13/SSC-CKA.png}
    %     \caption{VGG-13}
    %     \label{fig:vgg13c10b}
    %   \end{subfigure}
      \begin{subfigure}{.24\textwidth}
            \centering
            \includegraphics[width=\linewidth]{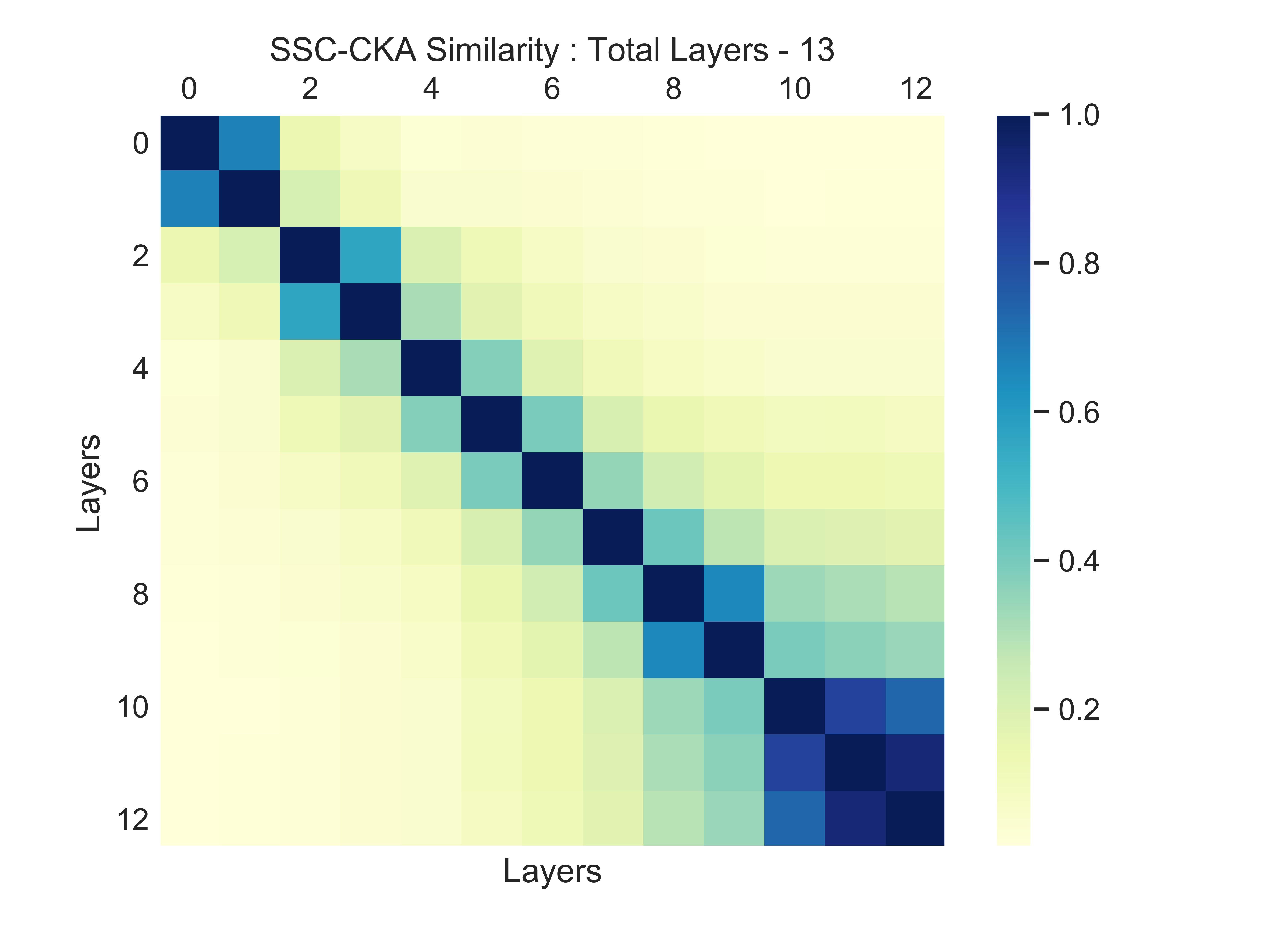}
            \caption{VGG-16}
            \label{fig:vgg16c10c}
          \end{subfigure}
        % \begin{subfigure}{.32\textwidth}
        %     \centering
        %     \includegraphics[width=\linewidth]{LayerWiseAnalysis/Cifar10/VGG19/SSC-CKA.png}
        %     \caption{VGG-19}
        %     \label{fig:vgg19c10d}
        %   \end{subfigure}
        \begin{subfigure}{.24\textwidth}
            \centering
            \includegraphics[width=\linewidth]{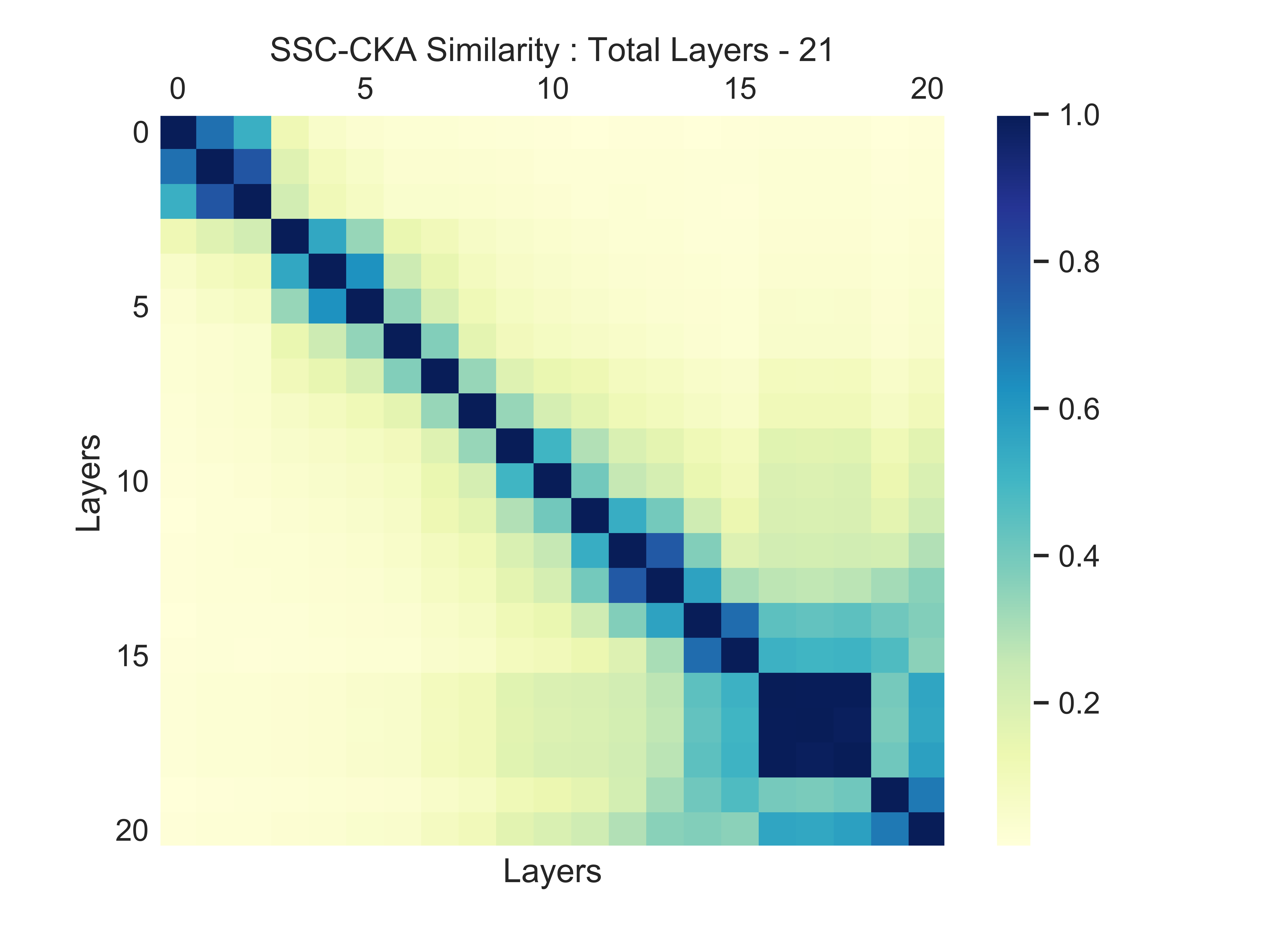}
            \caption{VGG-24}
            \label{fig:vgg24c10e}
          \end{subfigure}
         \begin{subfigure}{.24\textwidth}
            \centering
            \includegraphics[width=\linewidth]{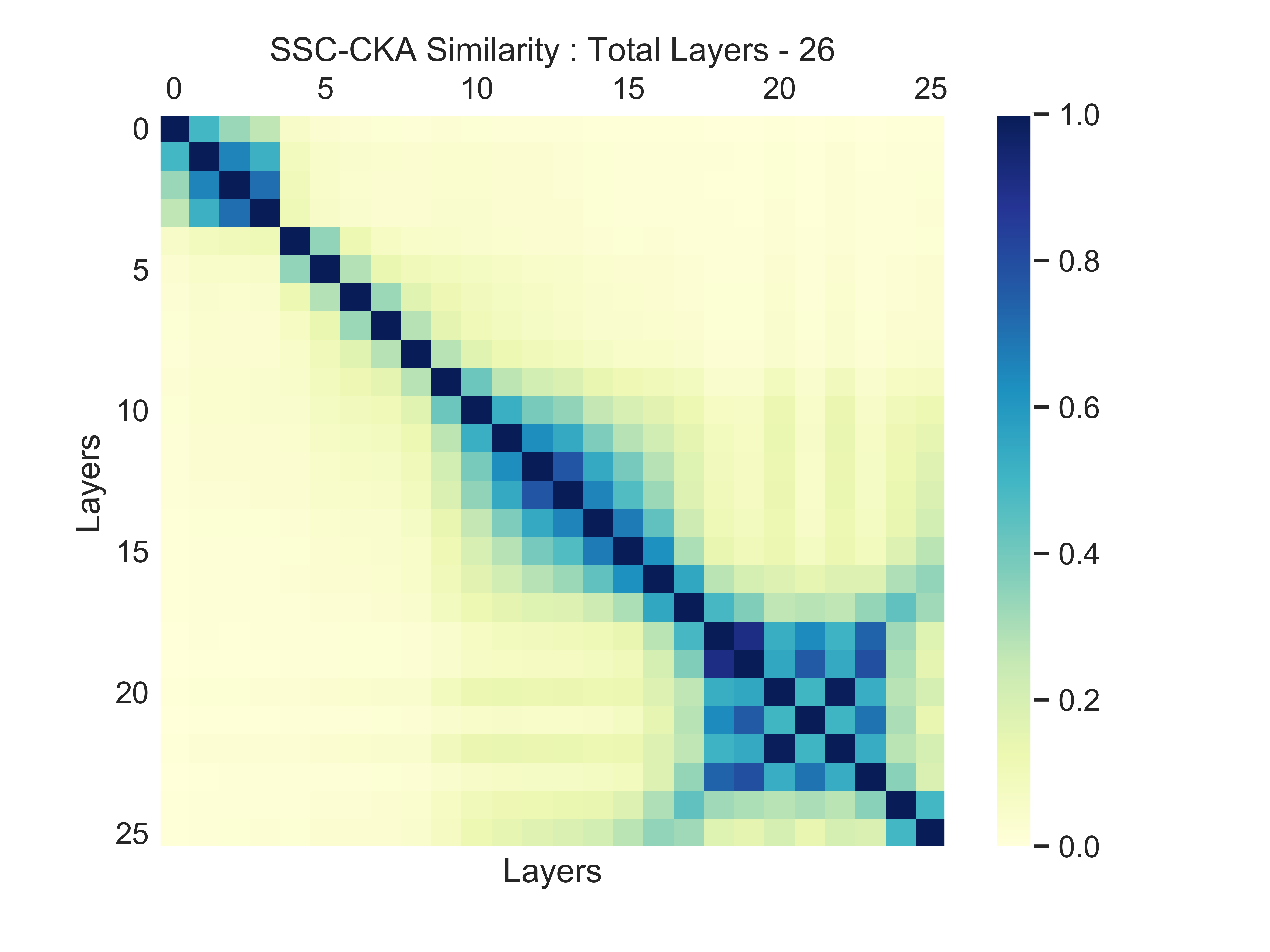}
            \caption{VGG-29}
            \label{fig:vgg29c10f}
          \end{subfigure}
    %   \caption{Left to Right : Pairwise SSC-CKA between all convolution layers of  VGG-11, VGG-13, VGG-16, VGG-19, VGG-24 and VGG-29 respectively - Data Set : CIFAR 10}
    \caption{Left to Right : Pairwise SSC-CKA between all convolution layers of  VGG-11: (accuracy 92\%), VGG-16: (accuracy 94\%), VGG-24: (accuracy 93\%) and VGG-29: (accuracy 62\%) respectively on CIFAR-10 dataset}. 
\end{figure*}

In this section we demonstrate the effect of depth on neural network architecture by analyzing the SSC-CKA maps of different CNNs with varying depths. Through  Figure \ref{fig:vgg11c10a} -  Figure \ref{fig:vgg29c10f}, we demonstrate the effects of increasing depths in VGG architecture based networks. In  Figure \ref{fig:vgg11c10a}, we train a VGG-11 and reach around 92\% accuracy on the evaluation set. Applying SSC-CKA to this network, we observe a prominently diagonal similarity matrix indicating that most layers, especially the earlier ones learn unique representations and as we go deeper in the architecture, some inter layer similarity appears. Upon increasing the depth in the architecture to a VGG-16 as shown in  Figure \ref{fig:vgg16c10c}, we observe a slight improvement in accuracy 94\%, but we begin to observe a diagonally dominant similarity matrix where shallower layers have some degree of similarity with their neighbours, but deeper layers, especially towards the end tend to form blocks of layers which are very similar to each other indicating that the network doesn't need to utilize the additional bandwidth to learn newer features in order to better learn and generalise. In  Figure \ref{fig:vgg24c10e}, we show a similar observation for VGG-24 that attained an accuracy of 93\%. However, as we increase the number of convolutional layers to 29 (VGG-29)  Figure \ref{fig:vgg29c10f}, the performance of the network drops to around 62\%, which is symptomatic of overfitting. We also observe a reinforcement in the prominence of the block diagonal structure signifying network over-parametrization relative to the data. 

% \subsubsection{Observing the effects of Depth}
% \begin{comment}
% \vspace{-.3cm}
\begin{figure*}[!ht]
% \begin{subfigure}{.32\textwidth}
%     \centering
%     \includegraphics[width=\linewidth]{LayerWiseAnalysis/Cifar10/ResNet18/SSC-CKA.png}
%     \caption{ResNet-18}
%     \label{fig:resnet18c10a}
%   \end{subfigure}
    \begin{subfigure}{.32\textwidth}
    \centering
    \includegraphics[width=\linewidth]{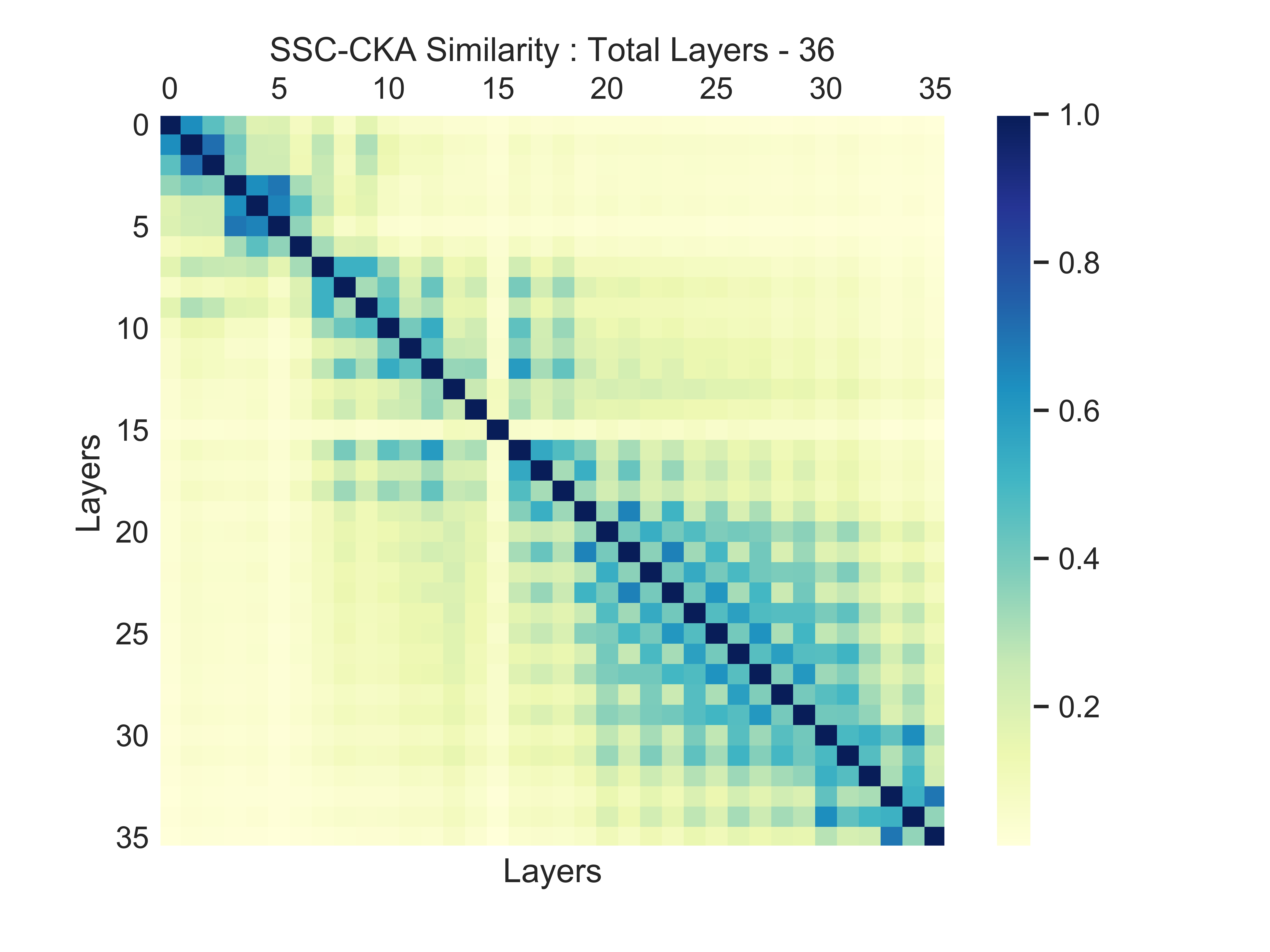}
    \caption{ResNet-36}
     \label{fig:resnet34c10b}
  \end{subfigure}
  \begin{subfigure}{.32\textwidth}
        \centering
        \includegraphics[width=\linewidth]{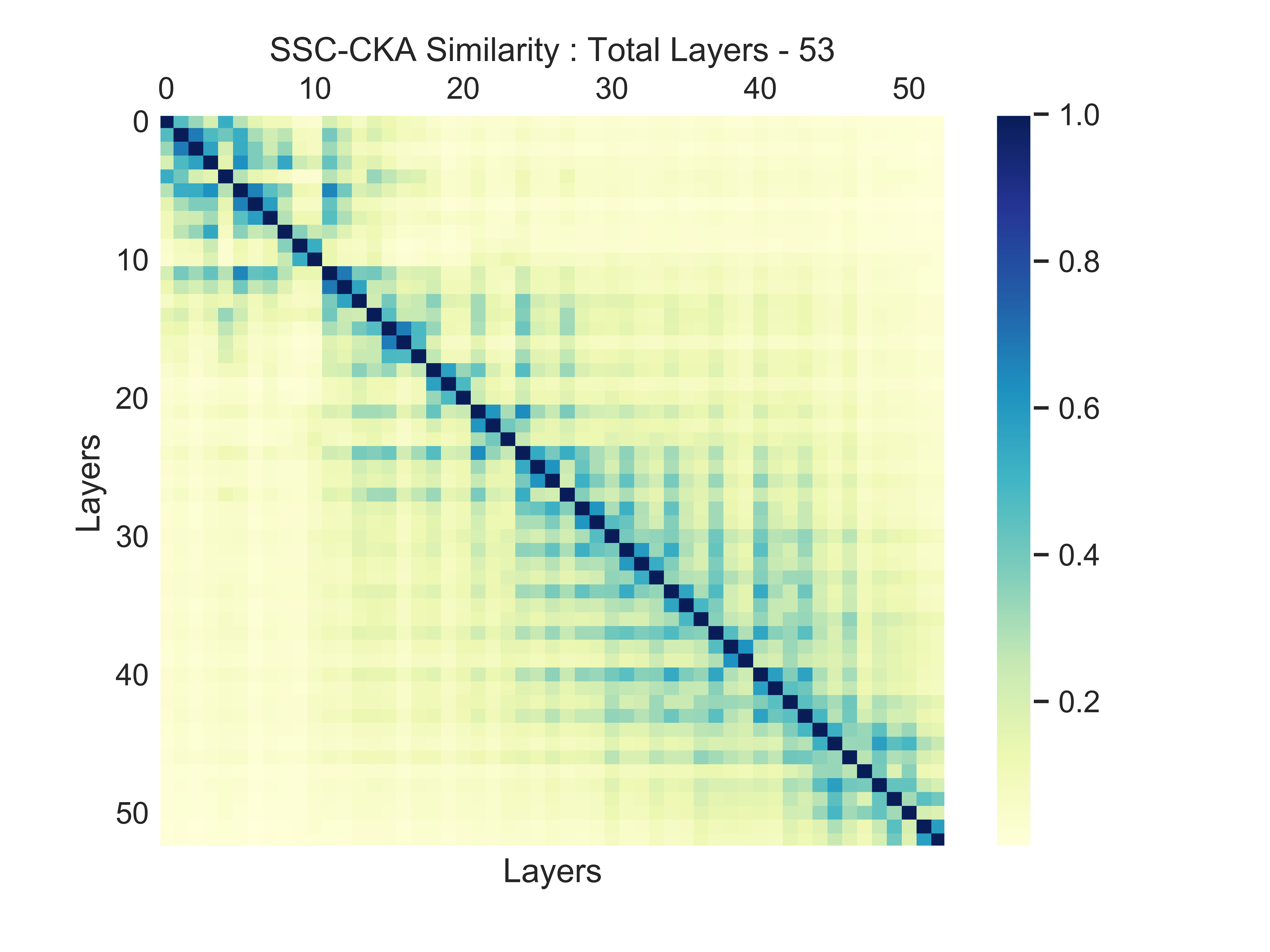}
        \caption{ResNet-53}
         \label{fig:resnet50c10c}
      \end{subfigure}
    \begin{subfigure}{.32\textwidth}
        \centering
        \includegraphics[width=\linewidth]{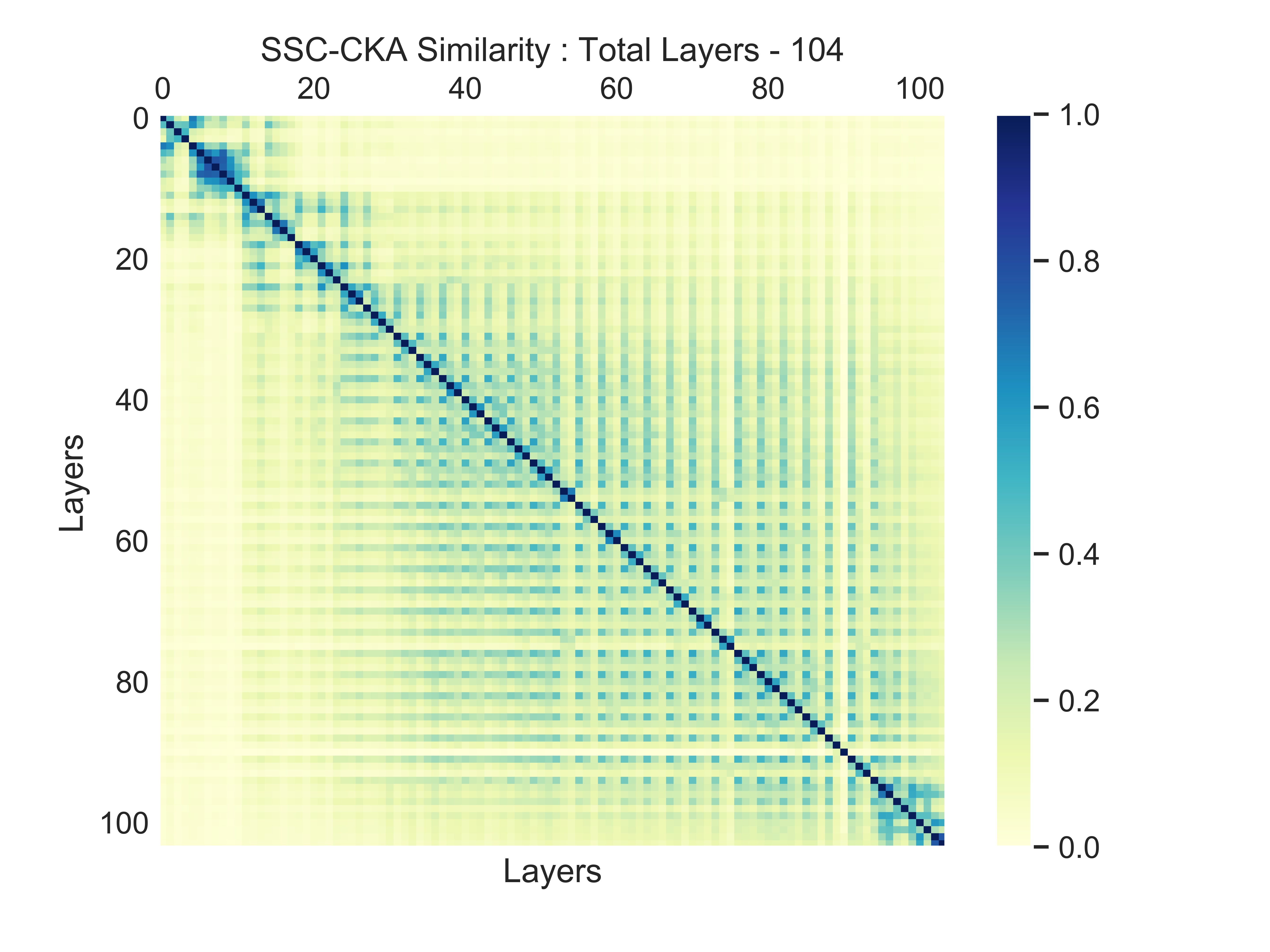}
        \caption{ResNet-104}
         \label{fig:resnet101c10d}
      \end{subfigure}
    % \begin{subfigure}{.45\textwidth}
    %     \centering
    %     \includegraphics[width=\linewidth]{LayerWiseAnalysis/Cifar10/ResNet152/SSC-CKA.png}
    %     \caption{ResNet-152}
    %      \label{fig:resnet152c10e}
    %   \end{subfigure}
    %   \caption{Left to Right : Pairwise SSC-CKA between all convolution layers of  ResNet-18, ResNet-36, ResNet-50, ResNet-101 and  ResNet-152, respectively - Data Set : CIFAR 10}
      \caption{Left to Right : Pairwise SSC-CKA between all convolution layers of   ResNet-36: (accuracy 94\%), ResNet-53: (accuracy 94\%), ResNet-104: (accuracy 93\%) respectively on CIFAR 10 dataset.}
\end{figure*}

% \vspace{-.6cm}
Next we demonstrate a similar set of results for ResNets of various depths,  Figure \ref{fig:resnet34c10b} -  Figure \ref{fig:resnet101c10d}. In case of ResNets, in addition to the block diagonal structure we also notice a chess-board pattern inside the blocks themselves, where every layer is similar to every alternate layer in a block. This is a consequence of  skip-connections present inside resnet blocks, which allow inputs from one layer to propagate much deeper into the network. The observations in this experiment are in-line with the ones made in \cite{kornblith2019similarity} and \cite{nguyen2020wide}.
% \vspace{-.3cm}

% \vspace{-.7cm}
\subsection{Observing the Effects of Width}
\begin{figure*}[th!]
 \begin{subfigure}{.32\textwidth}
    \centering
    \includegraphics[width=\linewidth]{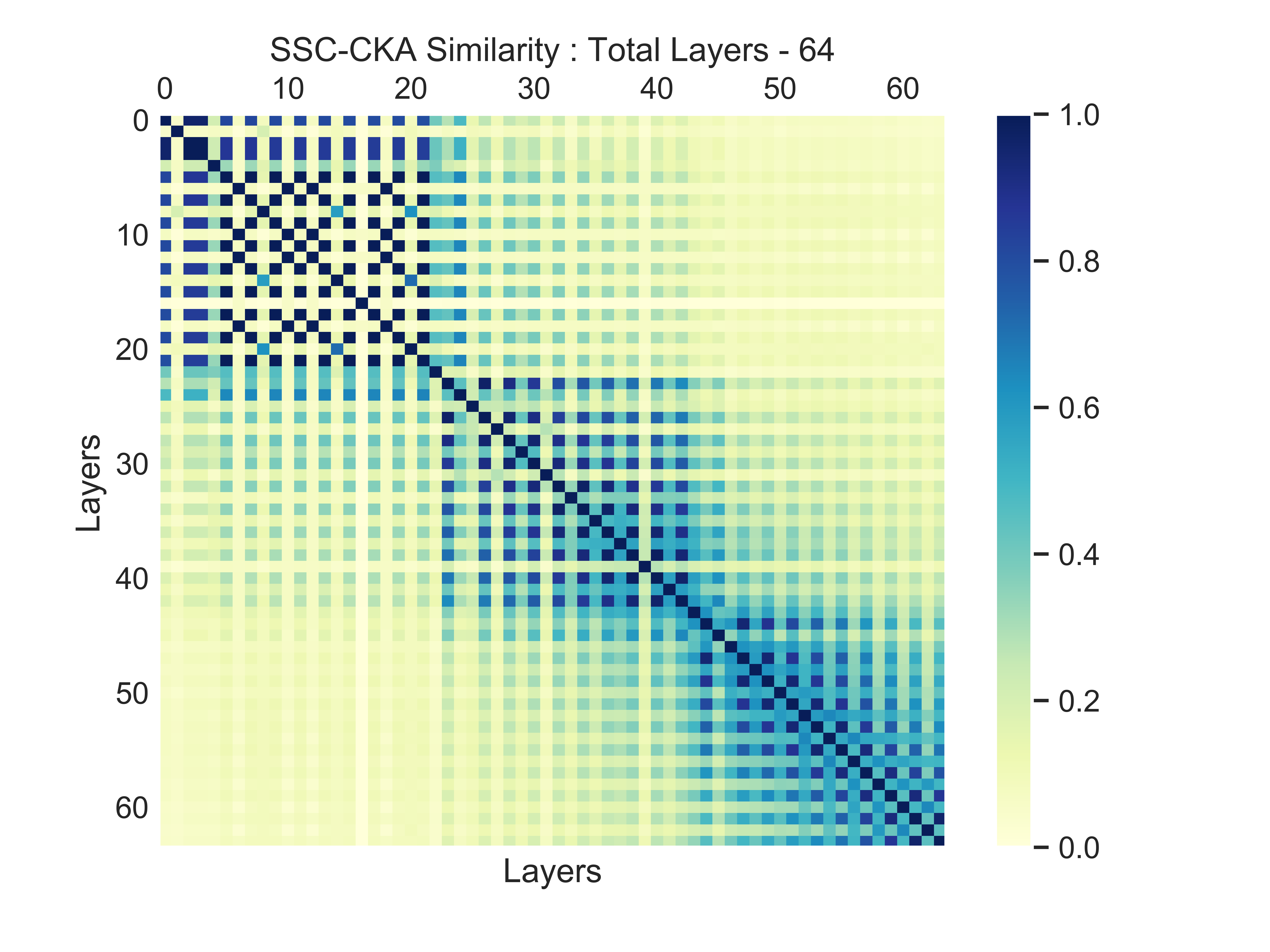}
    \caption{WR-64x2}
    \label{fig:wr64x2c10a}
  \end{subfigure}
%     \begin{subfigure}{.19\textwidth}
%     \centering
%     \includegraphics[width=\linewidth]{LayerWiseAnalysis/Cifar10/WRN64-4/SSC-CKA.png}
%     \caption{WR-64x4}
%     \label{fig:wr64x4c10b}
%   \end{subfigure}
  \begin{subfigure}{.32\textwidth}
        \centering
        \includegraphics[width=\linewidth]{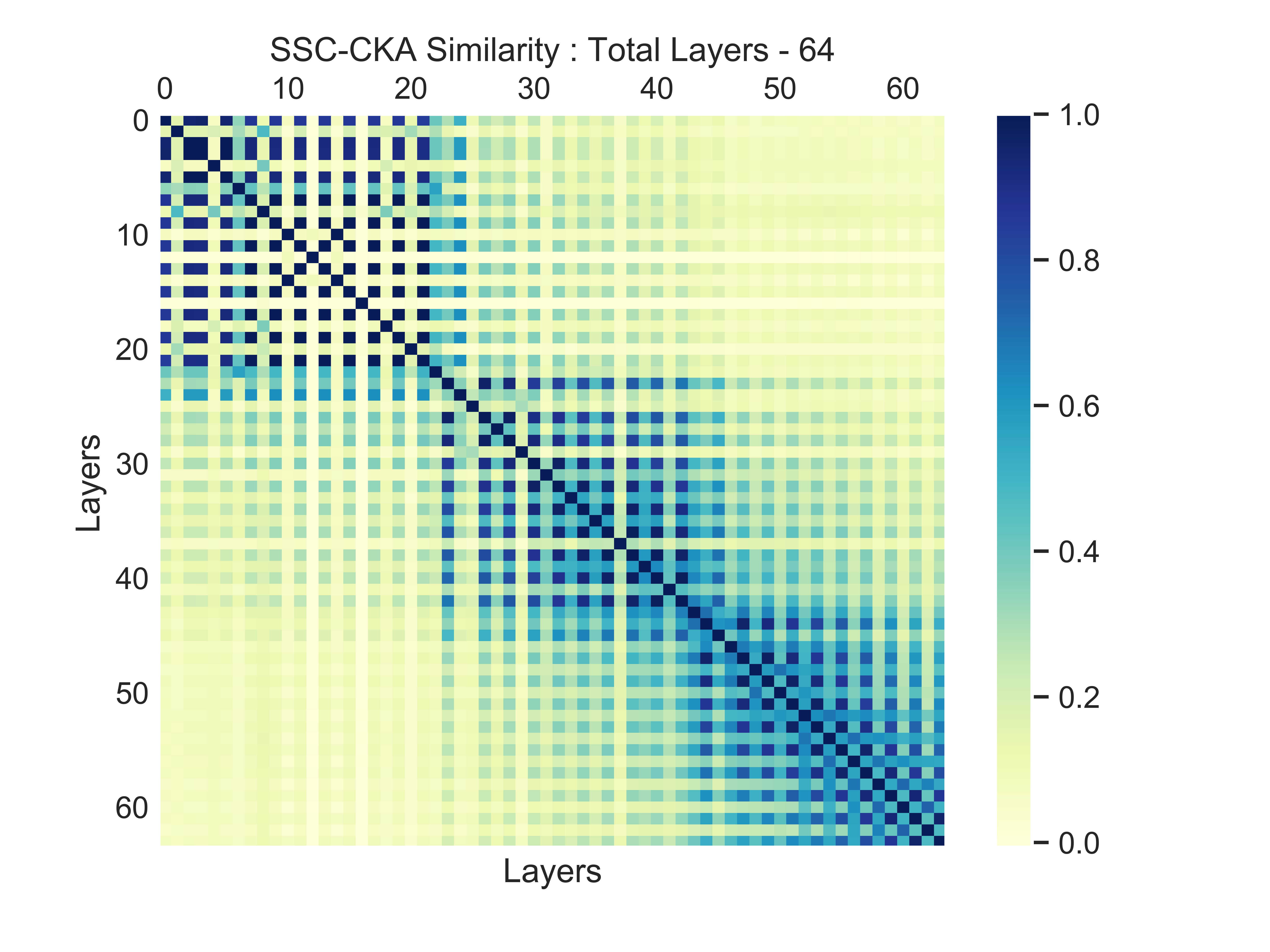}
        \caption{WR-64x6}
        \label{fig:wr64x6c10c}
      \end{subfigure}
    % \begin{subfigure}{.19\textwidth}
    %     \centering
    %     \includegraphics[width=\linewidth]{LayerWiseAnalysis/Cifar10/WRN64-8/SSC-CKA.png}
    %     \caption{WR-64x8}
    %     \label{fig:wr64x8c10d}
    %   \end{subfigure}
    \begin{subfigure}{.32\textwidth}
        \centering
        \includegraphics[width=\linewidth]{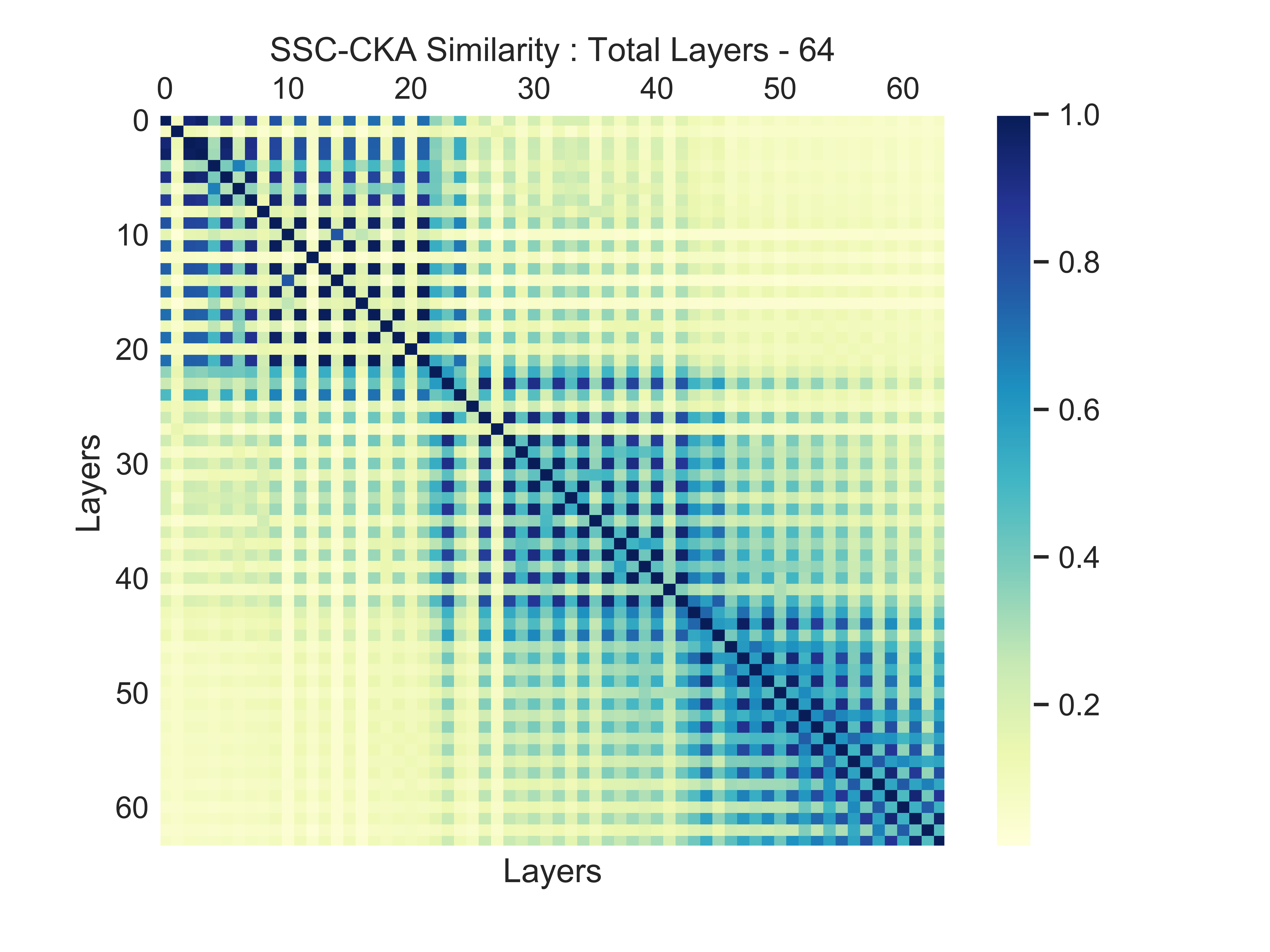}
        \caption{WR-64x10}
        \label{fig:wr64x10c10e}
      \end{subfigure}
    %   \caption{Left to Right : Pairwise SSC-CKA between all convolution layers of Wide ResNet-64 with width 2x, 4x, 6x, 8x and 10x, respectively - Data Set : CIFAR 10}
    \caption{Left to Right : Pairwise SSC-CKA between all convolution layers of Wide ResNet-64 with width 2x(Accuracy-95\%), 6x(Accuracy-95\%) and 10x(Accuracy-96\%), respectively on CIFAR-10 dataset.}
\end{figure*}

% \vspace{-.3cm}
In this experiment, we aim to study the effects of increasing the width of an architecture for a given depth. We use Wide-ResNet-64 network architecture with varying depth configurations (width 2x, 6x and 10x) and train them to their peak performance. The results of SSC-CKA are presented in  Figure \ref{fig:wr64x2c10a} -  Figure \ref{fig:wr64x10c10e}. In case of Wide-ResNets for a given depth, we observe that the architectural structure of various width configurations is very similar. The network can be divided into three distinct and disjoint blocks of layers, each having a high intra-block similarity and low inter-block similarity. Furthermore, we also notice the network wide presence of the chess-board pattern inherent to ResNets where even and odd layers in a layer block are dissimilar to each other, especially in the shallower layers of the network. These results present a contrast to those in \cite{nguyen2020wide}, as the authors there observe a marked increase in presence of a block structure as the width of a network increases.

% \vspace{-.7cm}
\subsection{Observing the Effects of Epochs}
\begin{figure*}[!ht]
 \begin{subfigure}{.24\textwidth}
    \centering
    \includegraphics[width=\linewidth]{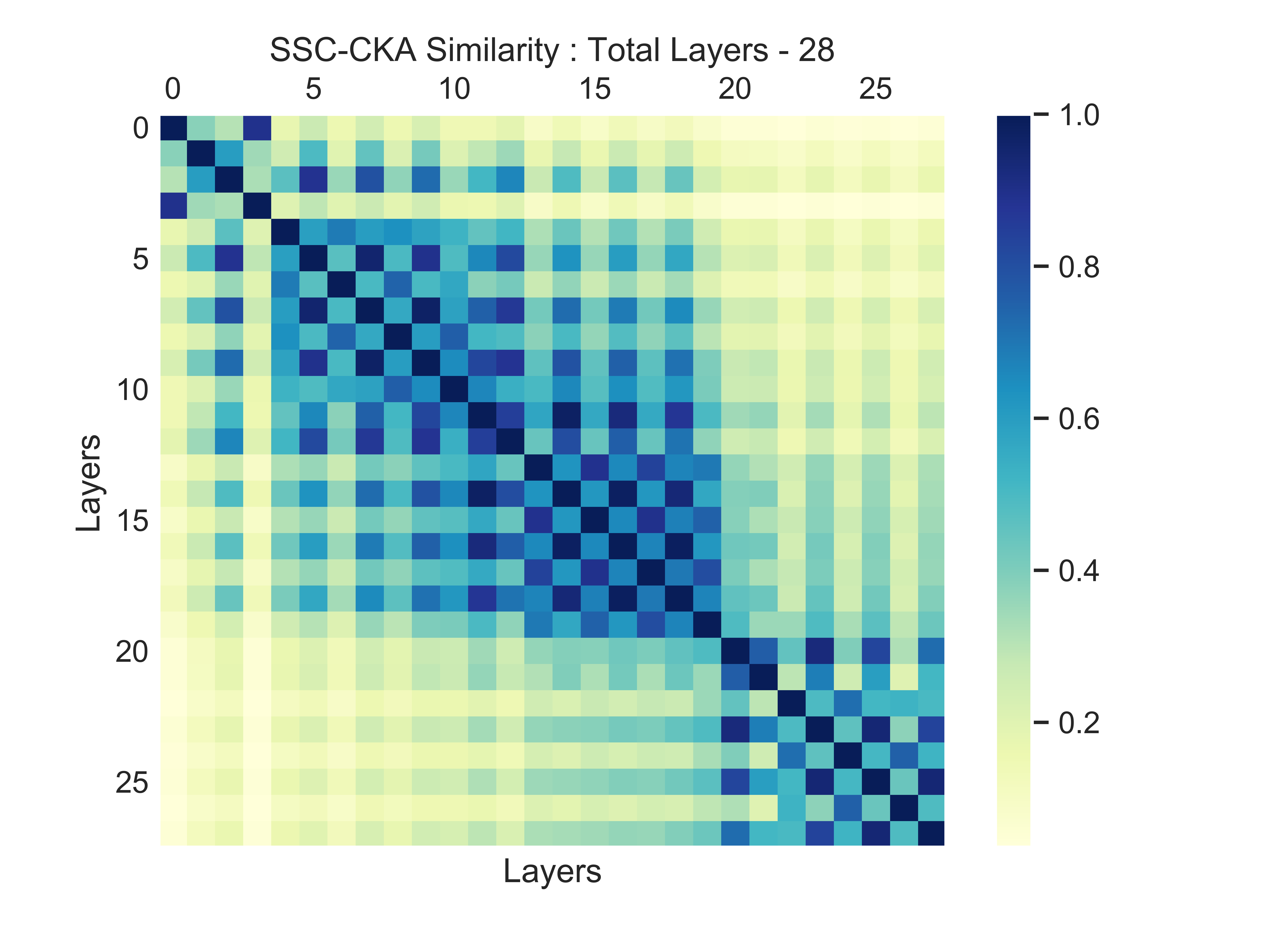}
    \caption{WR-28x2 After $1^{st}$ Epoch.}
    \label{fig:wr28x2c10e1}
  \end{subfigure}
    \begin{subfigure}{.24\textwidth}
    \centering
    \includegraphics[width=\linewidth]{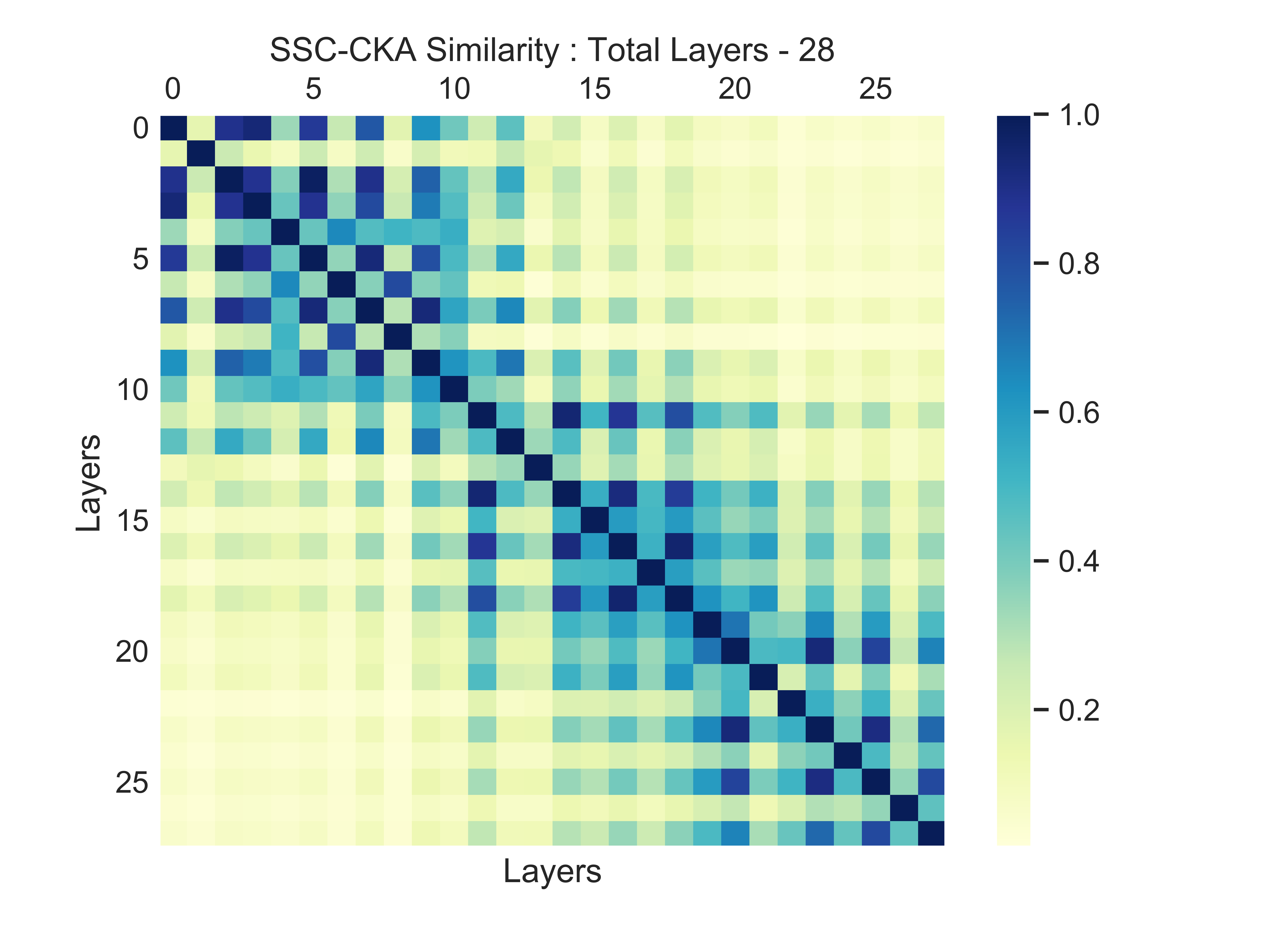}
    \caption{WR-28x2 After $31^{st}$ Epoch.}
    \label{fig:wr28x2c10e31}
  \end{subfigure}
  \begin{subfigure}{.24\textwidth}
        \centering
        \includegraphics[width=\linewidth]{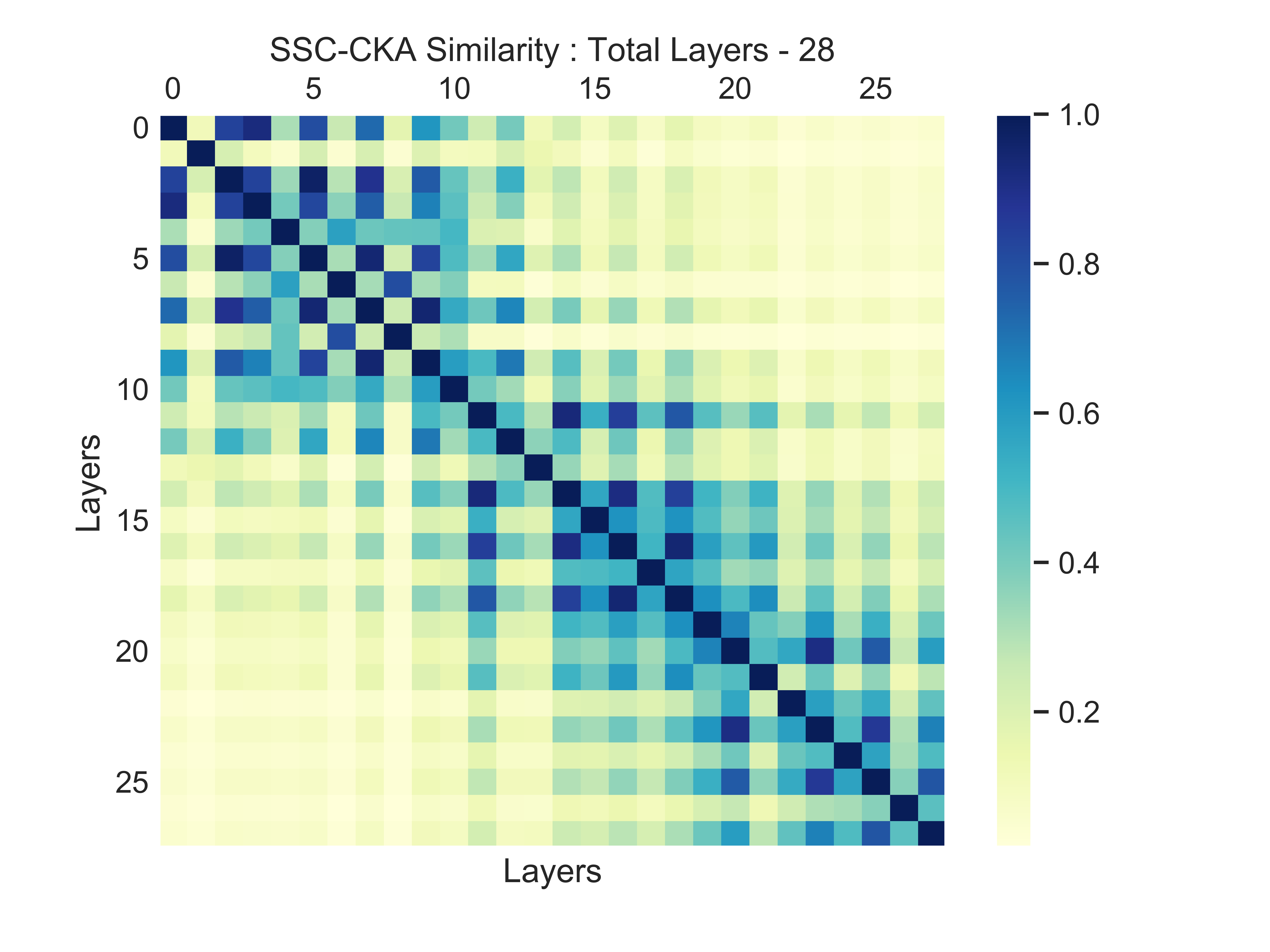}
        \caption{WR-28x2 After $61^{st}$ Epoch.}
        \label{fig:wr28x2c10e61}
      \end{subfigure}
    \begin{subfigure}{.24\textwidth}
        \centering
        \includegraphics[width=\linewidth]{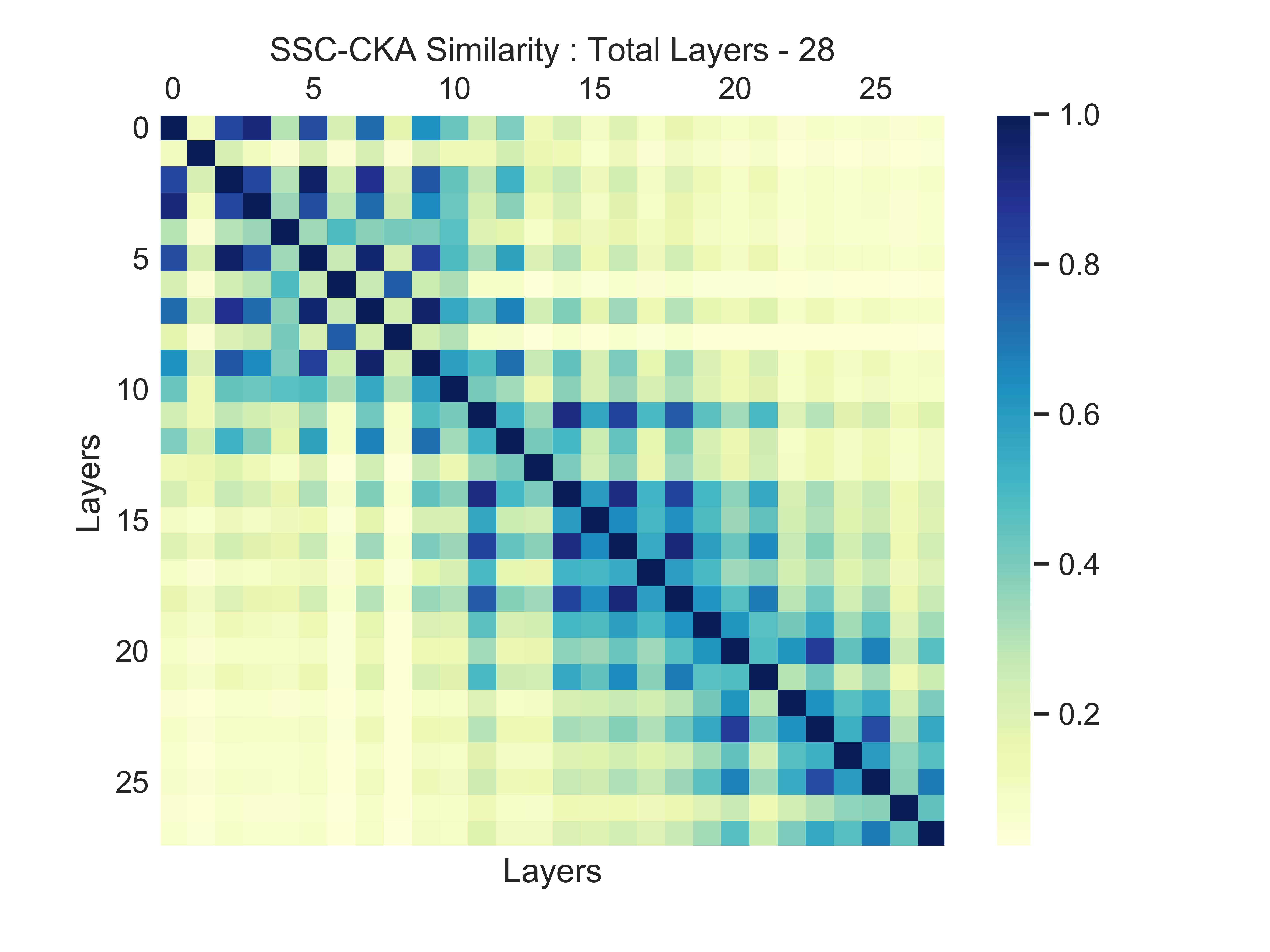}
        \caption{WR-28x2 After $100^{th}$ Epoch.}
        \label{fig:wr28x2c10e100}
      \end{subfigure}
      \caption{Left to Right : Pairwise SSC-CKA between all convolution layers of Various Wide ResNet-28x2 After $1^{st}$, $31^{st}$, $61^{st}$ and $100^{th}$ Epoch in Training on CIFAR-10 dataset.}
\end{figure*}

Here, we try to observe the behaviour of a network during the process of its training to better understand network training dynamics and visualize through SSC-CKA the process with which layers of a neural network organise themselves. We choose a Wide ResNet-28-by-2 and observe the SSC-CKA heatmaps after $1^{st}$ Epoch (35\% -  Figure \ref{fig:wr28x2c10e1}), $31^{st}$ Epoch (84\% -  Figure \ref{fig:wr28x2c10e31}), $61^{st}$ Epoch (91\% -  Figure \ref{fig:wr28x2c10e61}) and $100^{th}$ Epoch (95\% -  Figure \ref{fig:wr28x2c10e100}) of training where the network accuracy at that stage is indicated in parenthesis. After the first epoch when the network has an accuracy of around 35\% on the evaluation set, we observe an extremely large block diagonal structure encapsulating about the first two-third of the network. Subsequently as the network trains and improves its performance, the large global structure present in the earlier layers makes way for a smaller and more contained local block diagonal structures. This observation into network training dynamics can provide us with alternate ways to determine the maturation of the training process, one that doesn't necessarily need any labeled data. This experiment can be seen as an extension of the procedure demonstrated in  Figure \ref{fig:den121c10b} and  Figure \ref{fig:den121c100b}, but instead of comparing a few layers of the network per epoch to their final state, we compare all convolutional layers after every few epochs.

% \end{comment}
% \vspace{-.5cm}
\subsection{Observing the effects of Quantity of Training Data}
\begin{figure*}[!ht]
 \begin{subfigure}{.32\textwidth}
    \centering
    \includegraphics[width=\linewidth]{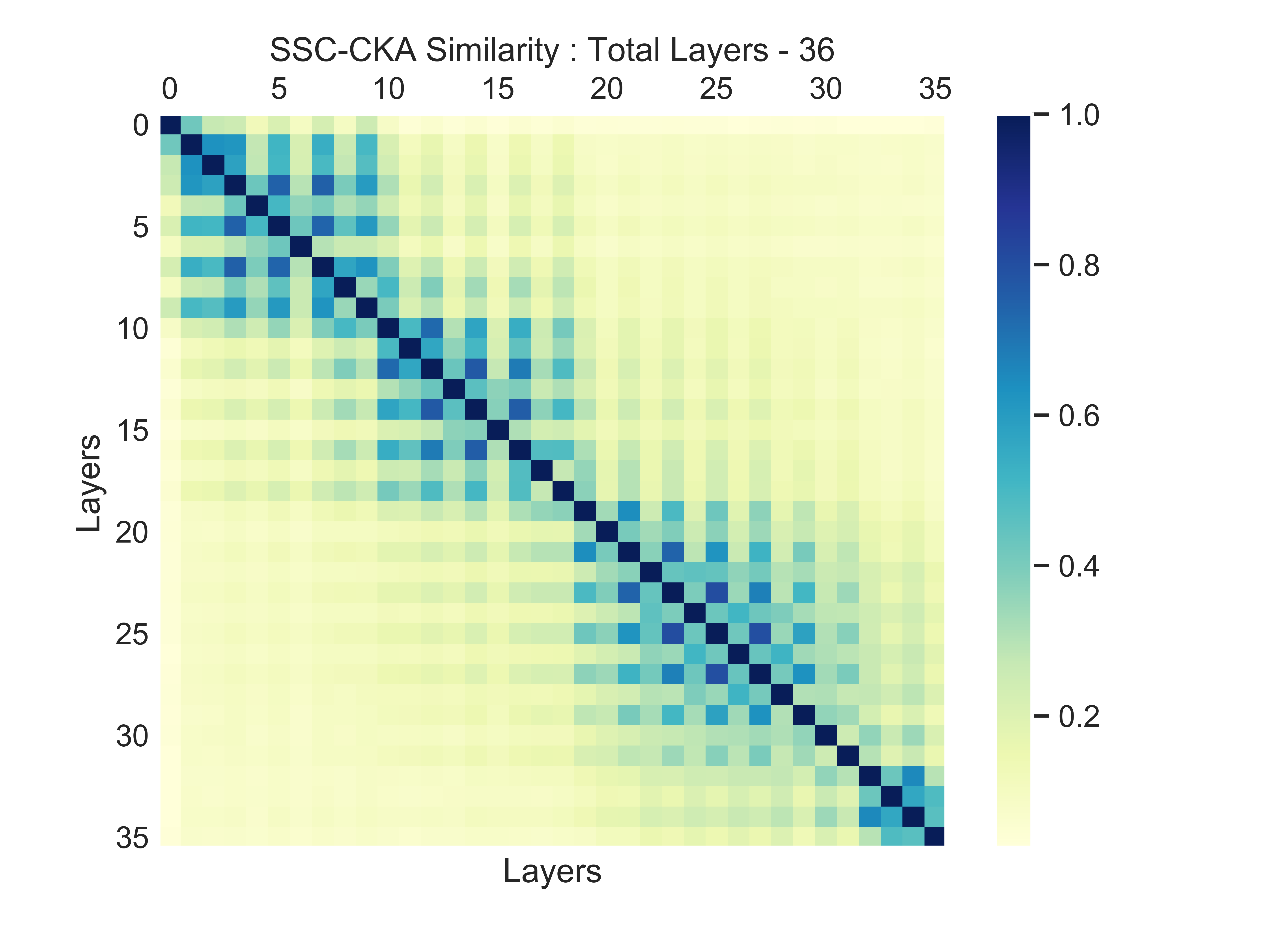}
    \caption{ResNet-36 - 5\% Training Data}
    \label{fig:ResNet36-5per}
  \end{subfigure}
    \begin{subfigure}{.32\textwidth}
    \centering
    \includegraphics[width=\linewidth]{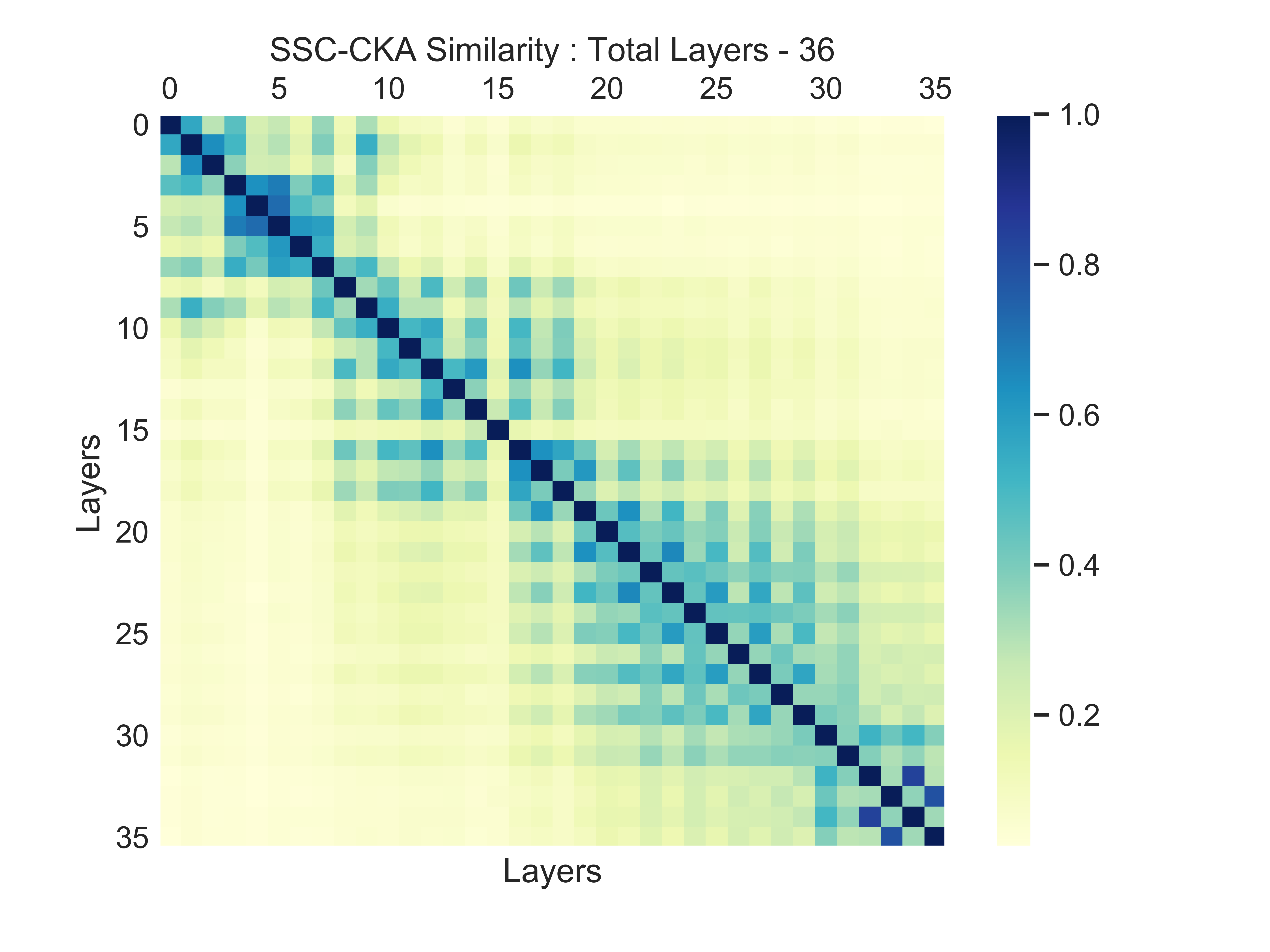}
    \caption{ResNet-36 - 50\% Training Data}
    \label{fig:ResNet36-50per}
  \end{subfigure}
  \begin{subfigure}{.32\textwidth}
        \centering
        \includegraphics[width=\linewidth]{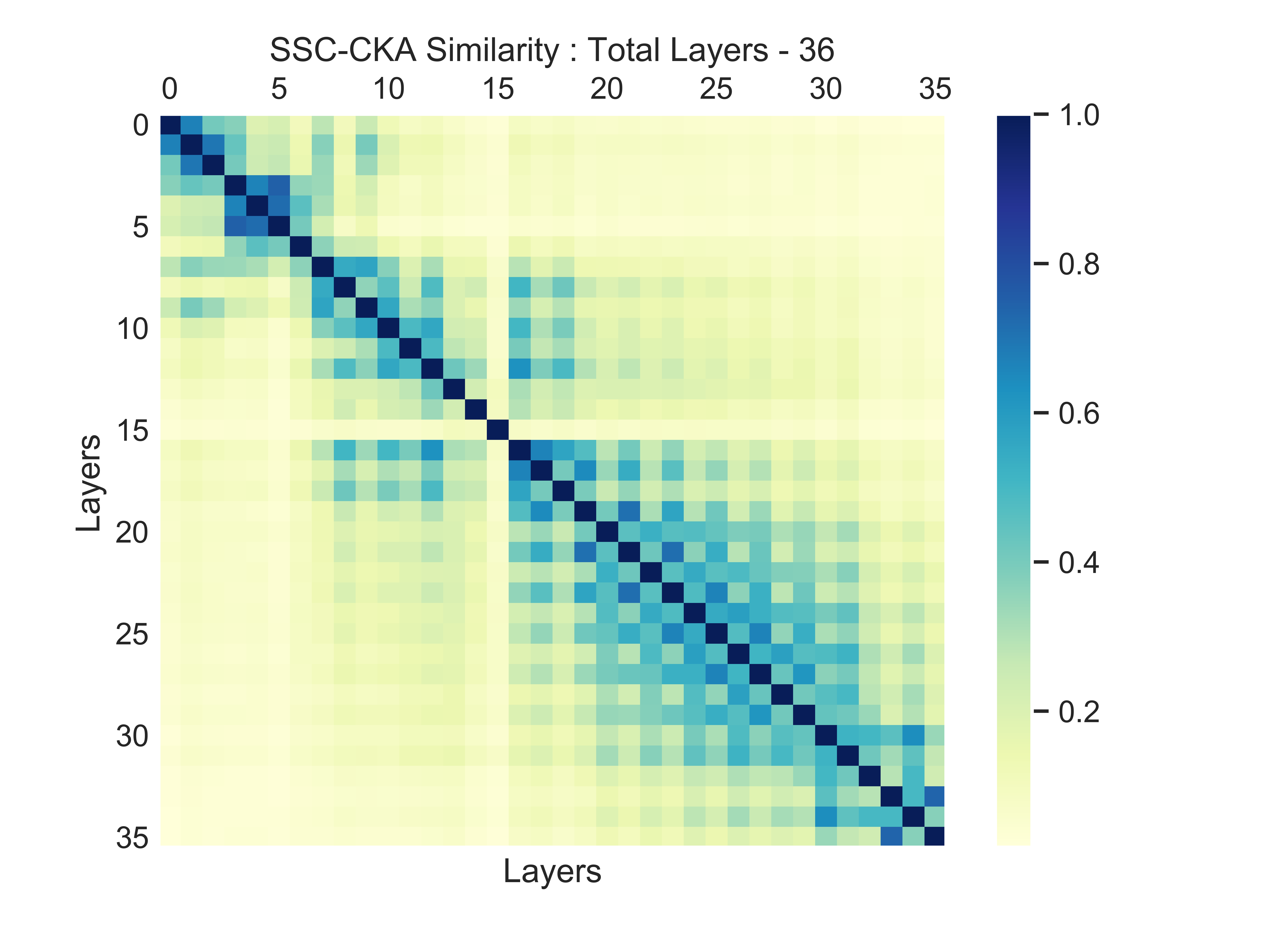}
        \caption{ResNet-36 - 100\% Training Data}
        \label{fig:ResNet36-100per}
      \end{subfigure}
      \caption{Left to Right : Pairwise SSC-CKA between all convolution layers of Various ResNet-36 when trained on 5\%, 50\% and 100\% Training Data on CIFAR-10 dataset.}
\end{figure*}

Next, along the lines of \cite{nguyen2020wide} we setup an experiment to demonstrate the emergence of the block diagonal structure in SSC-CKA heatmaps in networks that are over-parameterized relative to the training data. To simulate network size relative to data we train the network on 3 different dataset configurations (of CIFAR-10), the first configuration uses only 5\% of the original CIFAR-10 training set, the second uses 50\% of the training set and the third uses the entire set. The network used in these sets of experiments is a ResNet-36 trained till saturation of performance on the evaluation set. The network achieves an accuracy of 56\% when trained with only 5\% of the training data -  Figure \ref{fig:ResNet36-5per}, 92\% when trained with half the training set -  Figure \ref{fig:ResNet36-50per} and 94\% when trained with the entire training data -  Figure \ref{fig:ResNet36-100per}. Comparing the network that was trained on only 5\% of the dataset to the other two configurations we observe a more pronounced block-diagonal structure consisting of two distinct and disjointed blocks comprising the first half of the network in the former configuration. As the network is less starved for data, the block diagonal structure that was prevalent in the first half starts to become less recognisable until it breaks down into much more localised structures as seen in  Figure \ref{fig:ResNet36-100per}. These observations through SSC-CKA re-affirm the assertions made in Linear-CKA \cite{kornblith2019similarity}, \cite{nguyen2020wide} that over-parametrized networks are prone to developing a block diagonal structure as shown in their heat-maps for pairwise similarity. 
Please note that the second half of all the networks still retains a block diagonal structure. This is in accordance with the earlier observations made by increasing the depths of neural networks.
% \vspace{-.3cm}

\section{Analysis of Inputs}
\label{sec:resultsb}

In this section, we focus our analysis on individual inputs fed to the network as a part of the study. The goal is to highlight the ability of the model to interpret each input in terms of it's affinity to it's neighbours, and demonstrate experimentally that the neural network tries to separate out input of different classes into different mostly disjoint subspaces.

\subsection{Layer-wise Latent Space Visualisation}
\begin{figure*}[!ht]
 \begin{subfigure}{.32\textwidth}
    \centering
    \includegraphics[width=\linewidth]{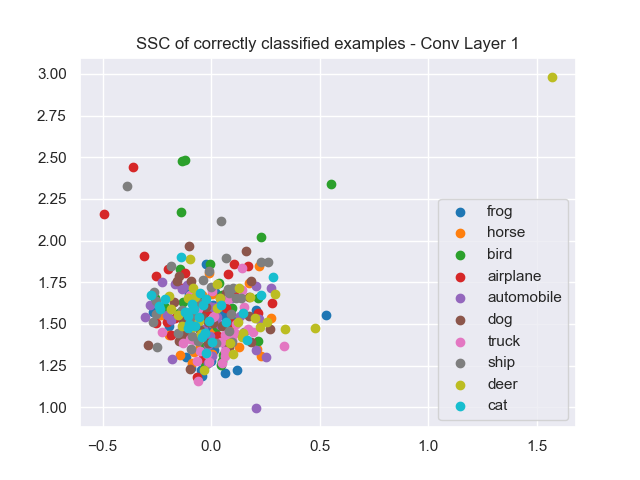}
    \caption{ResNet-36 : Layer 1}
    \label{fig:ResNet36c10-l1}
  \end{subfigure}
\begin{subfigure}{.32\textwidth}
    \centering
    \includegraphics[width=\linewidth]{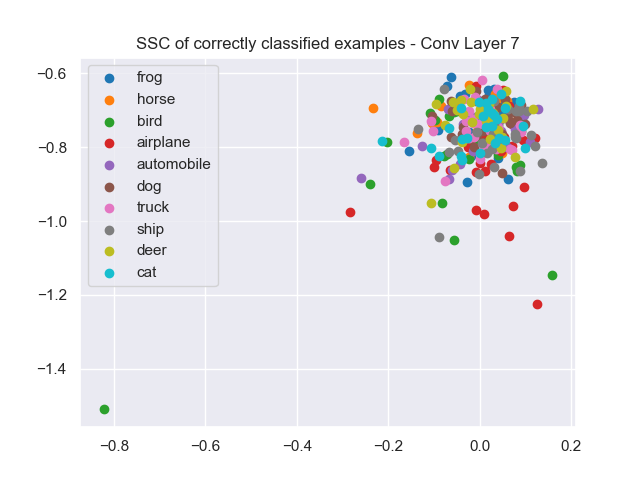}
    \caption{ResNet-36 : Layer 7}
    \label{fig:ResNet36c10-l7}
  \end{subfigure}
 \begin{subfigure}{.32\textwidth}
        \centering
        \includegraphics[width=\linewidth]{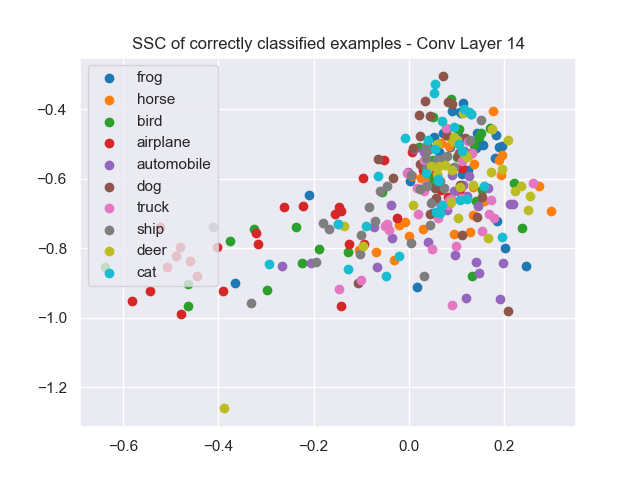}
        \caption{ResNet-36 : Layer 14}
        \label{fig:ResNet36c10-l14}
      \end{subfigure}
\begin{subfigure}{.32\textwidth}
        \centering
        \includegraphics[width=\linewidth]{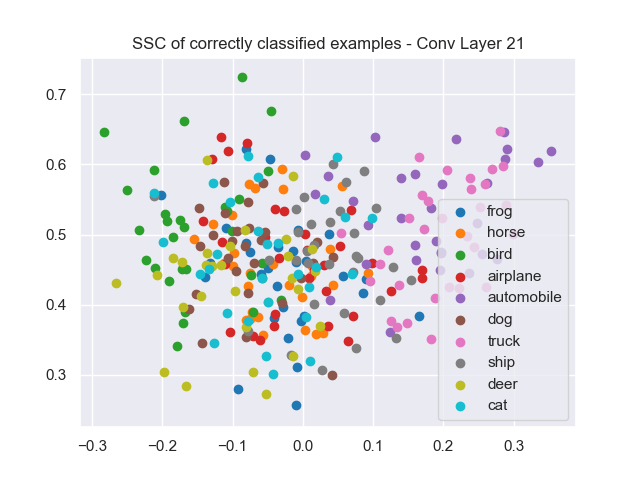}
        \caption{ResNet-36 : Layer 21}
        \label{fig:ResNet36c10-l21}
  \end{subfigure}
\begin{subfigure}{.32\textwidth}
    \centering
    \includegraphics[width=\linewidth]{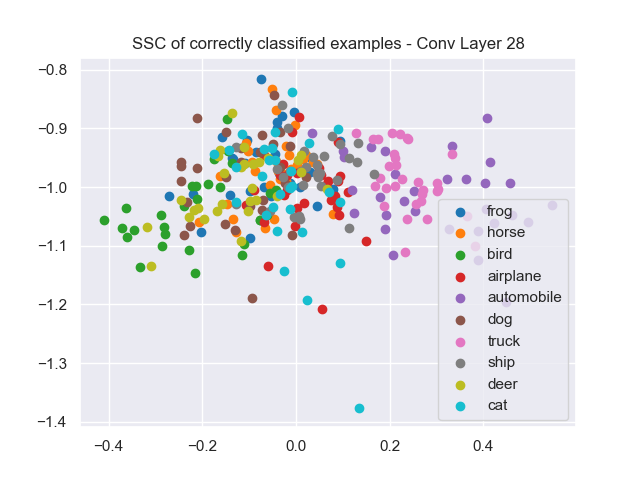}
    \caption{ResNet-36 : Layer 28}
    \label{fig:ResNet36c10-l28}
  \end{subfigure}
 \begin{subfigure}{.32\textwidth}
        \centering
        \includegraphics[width=\linewidth]{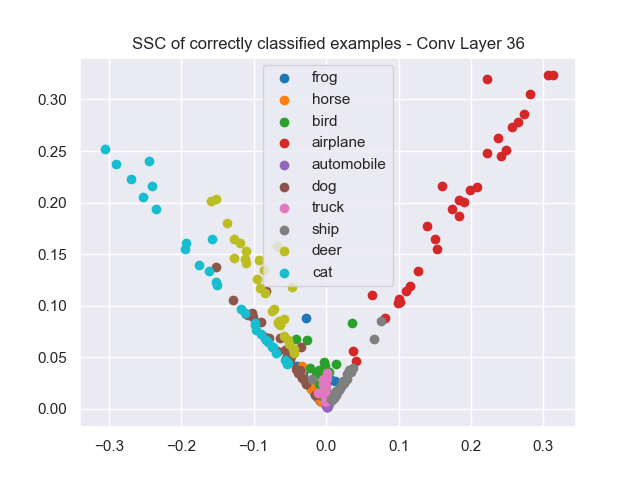}
        \caption{ResNet-36 : Layer 36}
        \label{fig:ResNet36c10-l36}
      \end{subfigure}
      \label{fig:ResNet36c10-SomeLayers}
      \caption{Left to Right : Spectral Embeddings of Layer-wise Affinity graphs learned via a ResNet-36 - Data Set : CIFAR 10}
\end{figure*}

To begin the analysis, we first present a selective layer-by-layer two-dimensional embedding of all correctly classified inputs to the network. We obtain the embeddings by taking the top two eigenvectors obtained from the decomposition of the respective layer's affinity matrix (Normalised Laplacian Matrix can also be used), which we do along the lines of \cite{NIPS2001_801272ee} and \cite{elhamifar2013sparse}. In  Figure \ref{fig:ResNet36c10-l1} -  Figure \ref{fig:ResNet36c10-l36}, we present the layer-wise analysis for six layers within the network, namely layer 1, 7, 14, 21, 28 and 36 respectively. In accordance with the observation made in  Figure \ref{sec:experiments} regarding the layer-by-layer modularity scores of the subspace affinity graph as we go deeper in the network, we observe a similar and perhaps a corroborating phenomenon, where inputs belonging to the same or similar classes get clustered closer to each other in a section of the latent space, and the inputs that belong to different classes get pushed into disjoint sections of the latent space, as we go deeper in the network.

% \vspace{-.4cm}
\subsection{Model Explanation by Instance Neighbourhood Visualisation}
Here, we analyze the failure cases and focus on the inputs where the network failed to classify the input correctly and chose inputs which had the highest output softmax scores among the incorrectly classified inputs, i.e. the network was confidently wrong for those inputs. For the purpose of this subsequent analysis, we take the SSC-based affinity of the network's final convolution layer.

\begin{figure*}[!h]
\centering
 \begin{subfigure}{.2\textwidth}
    \centering
    \includegraphics[width=\linewidth]{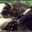}
    \caption{Input}
    \label{fig:catTobird1}
  \end{subfigure}
\begin{subfigure}{.75\textwidth}
    \centering
    \includegraphics[width=\linewidth]{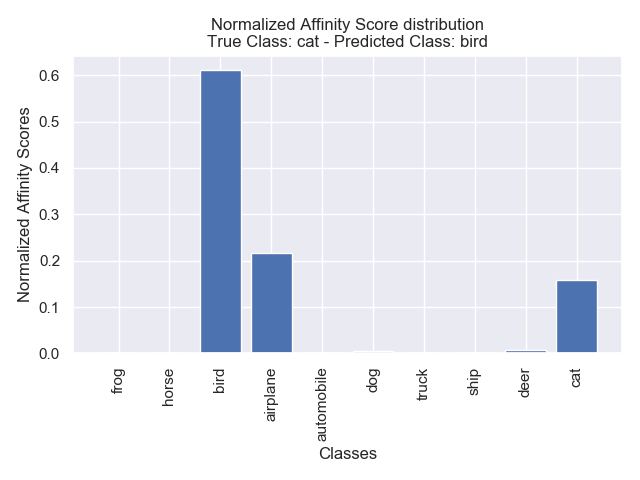}
    \caption{Neighbourhood Distribution (Normalised)}
    \label{fig:AffScores1_1}
  \end{subfigure}

  \centering
 \begin{subfigure}{.1\textwidth}
        \centering
        \includegraphics[width=\linewidth]{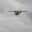}
        % \caption{Airplane - 0.055}
        \caption{Airplane}
        \label{fig:n1_1}
      \end{subfigure}
\begin{subfigure}{.1\textwidth}
        \centering
        \includegraphics[width=\linewidth]{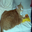}
        % \caption{Cat - 0.052}
        \caption{Cat}
        \label{fig:n1_2}
  \end{subfigure}
\begin{subfigure}{.1\textwidth}
    \centering
    \includegraphics[width=\linewidth]{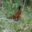}
    % \caption{Bird - 0.044}
    \caption{Bird}
    \label{fig:n1_3}
  \end{subfigure}
 \begin{subfigure}{.1\textwidth}
        \centering
        \includegraphics[width=\linewidth]{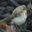}
        % \caption{Bird - 0.04}
        \caption{Bird}
        \label{fig:n1_4}
      \end{subfigure}
      \begin{subfigure}{.1\textwidth}
        \centering
        \includegraphics[width=\linewidth]{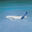}
        % \caption{Airplane - 0.038}
        \caption{Airplane}
        \label{fig:n1_5}
      \end{subfigure}
      \begin{subfigure}{.1\textwidth}
        \centering
        \includegraphics[width=\linewidth]{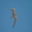}
        % \caption{Bird - 0.038}
        \caption{Bird}
        \label{fig:n1_6}
      \end{subfigure}
      \begin{subfigure}{.1\textwidth}
        \centering
        \includegraphics[width=\linewidth]{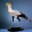}
        % \caption{Bird - 0.025}
        \caption{Bird}
        \label{fig:n1_7}
      \end{subfigure}
      \begin{subfigure}{.1\textwidth}
        \centering
        \includegraphics[width=\linewidth]{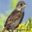}
        % \caption{Bird - 0.022}
        \caption{Bird}
        \label{fig:n1_8}
      \end{subfigure}
      \label{fig:n1}
      \caption{9(a): A Cat, classified as a bird with a softmax score of 0.97. 9(b): Distribution of it's Neighbourhood Affinity Scores (Normalised). 9(c): Airplane. 9(d): Cat. 9(e): Bird. 9(f): Bird. 9(g): Airplane. 9(h): Bird. 9(i): Bird. 9(j): Bird. Network: ResNet-36 - Data Set : CIFAR 10}
\end{figure*}

In   Figure \ref{fig:catTobird1} we show the input image of a cat that was incorrectly classified as a bird with a softmax score of 0.97.  Figure \ref{fig:AffScores1_1} shows the normalised class-wise distribution of affinity scores for this input summed over all the images. This plot clearly shows the three strongest classes for which the given input has a strong affinity for, namely, bird, airplane and cat, in that order. At this juncture, we make a note that the predicted label, though incorrectly, of this input by the network seems to be the class for which this input has the highest affinity for as determined by Sparse Subspace Clustering Algorithm.   Figure \ref{fig:n1_1}- Figure \ref{fig:n1_8} show the top-8 images, in descending order, for which the given input of cat has the highest affinity for.

Continuing discussion on the previous observation, we further expand on that observation in   Table \ref{table:1}, where we present the accuracy of the network on the testing set in two scenarios. The first scenario assigns an output label to a given input based on the highest aggregate class-wise affinity score for that input and the second scenario assigns the same output label to the input as the networks prediction through it's classification layers. The results of those are presented in columns 2 and 3 labeled `SSC Label' and `Network Prediction' respectively, in  Table \ref{table:1}. We observe that both these `accuracies' turn out to be 95.8\%, Please be advised that the equality of values is a coincidence. Thus, given the activations of the final convolutional layer for the entire testing set we observe that the SSC based Affinity scoring assigns labels which are competitive with what the classification layers of the network could learn. This is further demonstrated in column 4 of   Table \ref{table:1} which shows a 98.3\% agreement between one to one comparisons of network assigned labels with SSC derived labels. Row 3 and row 4 show the same metrics for other 2 networks, on CIFAR-10. Such a high annotation agreement between SSC and network prediction indicates a tendency of neural networks to separate out data points belonging to different classes into disjoint sections of the latent embedding space.

% \vspace{-.5cm}
\begin{table}[ht!]
\centering
\caption{\textbf{SSC Labels vs. Network Labels}: Comparison of Accuracy and Correlation on the test set}
\begin{tabular}{ |p{3cm}||p{3cm}|p{3cm}|p{3cm}|  }
 \hline
 \multicolumn{4}{|c|}{Accuracy of Labels when compared to ground truth (CIFAR-10)} \\
 \hline
  Network & SSC Label & Network Prediction & Correlation \\
 \hline
  ResNet-34  & 95.8\%    & 95.8\% & 98.3\% \\
 \hline
 ResNet-18  & 95.4\%    & 96.1\% & 97\% \\
 \hline
 ResNet-50  & 90.6\%    & 94.8\% & 93.5\% \\
 \hline
\end{tabular}
% \bigskip
 
\label{table:1}
\end{table}

% \section{Related Work}
% \label{sec:related}
% \input{06_related}

\section{Conclusions}
\label{sec:conclusion}
Our work proposes a two-step framework to analyze deep neural networks. This framework combines Sparse Subspace Clustering and Centered Kernel Alignment to provide the ability to analyze the network on a macro and micro level. The macro analysis helps in visualising network architectures as a function of network depth, width, training epochs and training data quantity and provides an insight into network architecture and training dynamics of the network. It also provides a framework to compare two architecturally different neural network based on a common set of inputs. The framework also provides the ability to micro analyze the network in the form of instance based interpretability by providing a measure of a degree of closeness each input has to decision boundaries for different classes in the loss landscape of a network.

\section*{Acknowledgements}
{%\scriptsize
This research was supported by National Science Foundation Grant No. 1901379, the U.S. Department of Energy, Office of Science, Office of Advanced Scientific Computing Research (ASCR), Scientific Discovery through Advanced Computing (SciDAC) program, specifically the RAPIDS-2 SciDAC institute. Any opinions, findings, and conclusions or recommendations expressed in this material are those of the author(s) and do not necessarily reflect the views of the funding parties. The authors would like to thank Tushar Nagarajan for helpful discussions and anonymous reviewers for critical feedback.
}
%}

\bibliographystyle{splncs04}
% \bibliography{bib/refs.bib}

\begin{thebibliography}{10}
\providecommand{\url}[1]{\texttt{#1}}
\providecommand{\urlprefix}{URL }
\providecommand{\doi}[1]{https://doi.org/#1}

\bibitem{10.1016/S0304-3975(97)00115-1}
Amaldi, E., Kann, V.: On the approximability of minimizing nonzero variables or
  unsatisfied relations in linear systems. Theor. Comput. Sci.
  \textbf{209}(1–2),  237–260 (Dec 1998).
  \doi{10.1016/S0304-3975(97)00115-1},
  \url{https://doi.org/10.1016/S0304-3975(97)00115-1}

\bibitem{bau2017network}
Bau, D., Zhou, B., Khosla, A., Oliva, A., Torralba, A.: Network dissection:
  Quantifying interpretability of deep visual representations (2017)

\bibitem{10.1561/2200000016}
Boyd, S., Parikh, N., Chu, E., Peleato, B., Eckstein, J.: Distributed
  optimization and statistical learning via the alternating direction method of
  multipliers. Found. Trends Mach. Learn.  \textbf{3}(1),  1–122 (Jan 2011).
  \doi{10.1561/2200000016}, \url{https://doi.org/10.1561/2200000016}

\bibitem{DBLP:journals/corr/abs-1203-0550}
Cortes, C., Mohri, M., Rostamizadeh, A.: Algorithms for learning kernels based
  on centered alignment. CoRR  \textbf{abs/1203.0550} (2012),
  \url{http://arxiv.org/abs/1203.0550}

\bibitem{elhamifar2013sparse}
Elhamifar, E., Vidal, R.: Sparse subspace clustering: Algorithm, theory, and
  applications. IEEE transactions on pattern analysis and machine intelligence
  \textbf{35}(11),  2765--2781 (2013)

\bibitem{10.1007/11564089_7}
Gretton, A., Bousquet, O., Smola, A., Scholkopf, B.: Measuring statistical
  dependence with hilbert-schmidt norms. In: Proceedings of the 16th
  International Conference on Algorithmic Learning Theory. p. 63–77. ALT'05,
  Springer-Verlag, Berlin, Heidelberg (2005). \doi{10.1007/11564089_7},
  \url{https://doi.org/10.1007/11564089_7}

\bibitem{hanin2018approximating}
Hanin, B., Sellke, M.: Approximating continuous functions by relu nets of
  minimal width (2018)

\bibitem{he2015deep}
He, K., Zhang, X., Ren, S., Sun, J.: Deep residual learning for image
  recognition (2015)

\bibitem{huang2018densely}
Huang, G., Liu, Z., van~der Maaten, L., Weinberger, K.Q.: Densely connected
  convolutional networks (2018)

\bibitem{kornblith2019similarity}
Kornblith, S., Norouzi, M., Lee, H., Hinton, G.: Similarity of neural network
  representations revisited (2019)

\bibitem{Krizhevsky09learningmultiple}
Krizhevsky, A.: Learning multiple layers of features from tiny images. Tech.
  rep. (2009)

\bibitem{lee2018deep}
Lee, J., Bahri, Y., Novak, R., Schoenholz, S.S., Pennington, J.,
  Sohl-Dickstein, J.: Deep neural networks as gaussian processes (2018)

\bibitem{lin2018resnet}
Lin, H., Jegelka, S.: Resnet with one-neuron hidden layers is a universal
  approximator (2018)

\bibitem{lu2017expressive}
Lu, Z., Pu, H., Wang, F., Hu, Z., Wang, L.: The expressive power of neural
  networks: A view from the width (2017)

\bibitem{morcos2018insights}
Morcos, A.S., Raghu, M., Bengio, S.: Insights on representational similarity in
  neural networks with canonical correlation (2018)

\bibitem{Newman8577}
Newman, M.E.J.: Modularity and community structure in networks. Proceedings of
  the National Academy of Sciences  \textbf{103}(23),  8577--8582 (2006).
  \doi{10.1073/pnas.0601602103}, \url{https://www.pnas.org/content/103/23/8577}

\bibitem{NIPS2001_801272ee}
Ng, A., Jordan, M., Weiss, Y.: On spectral clustering: Analysis and an
  algorithm. In: Dietterich, T., Becker, S., Ghahramani, Z. (eds.) Advances in
  Neural Information Processing Systems. vol.~14. MIT Press (2002),
  \url{https://proceedings.neurips.cc/paper/2001/file/801272ee79cfde7fa5960571fee36b9b-Paper.pdf}

\bibitem{nguyen2020wide}
Nguyen, T., Raghu, M., Kornblith, S.: Do wide and deep networks learn the same
  things? uncovering how neural network representations vary with width and
  depth (2020)

\bibitem{novak2020bayesian}
Novak, R., Xiao, L., Lee, J., Bahri, Y., Yang, G., Hron, J., Abolafia, D.A.,
  Pennington, J., Sohl-Dickstein, J.: Bayesian deep convolutional networks with
  many channels are gaussian processes (2020)

\bibitem{raghu2017svcca}
Raghu, M., Gilmer, J., Yosinski, J., Sohl-Dickstein, J.: Svcca: Singular vector
  canonical correlation analysis for deep learning dynamics and
  interpretability (2017)

\bibitem{simonyan2014deep}
Simonyan, K., Vedaldi, A., Zisserman, A.: Deep inside convolutional networks:
  Visualising image classification models and saliency maps (2014)

\bibitem{simonyan2015deep}
Simonyan, K., Zisserman, A.: Very deep convolutional networks for large-scale
  image recognition (2015)

\bibitem{10.5555/2948884}
Vidal, R., Ma, Y., Sastry, S.S.: Generalized Principal Component Analysis.
  Springer Publishing Company, Incorporated, 1st edn. (2016)

\bibitem{zagoruyko2017wide}
Zagoruyko, S., Komodakis, N.: Wide residual networks (2017)

\bibitem{zeiler2013visualizing}
Zeiler, M.D., Fergus, R.: Visualizing and understanding convolutional networks
  (2013)

\end{thebibliography}

\end{document}